%% file: main.tex
\documentclass{article}
\usepackage[preprint, nonatbib]{neurips_2025}
\usepackage[utf8]{inputenc}     % allow utf-8 input
\usepackage[T1]{fontenc}        % use 8-bit T1 fonts
\usepackage[dvipsnames]{xcolor} % colors
\usepackage{hyperref}           % hyperlinks
\usepackage{url}                % simple URL typesetting
\usepackage{booktabs}           % professional-quality tables
\usepackage{amsfonts}           % blackboard math symbols
\usepackage{nicefrac}           % compact symbols for 1/2, etc.
\usepackage{microtype}          % microtypography
\usepackage{tcolorbox}
\tcbuselibrary{skins}
\usepackage{graphicx}
\usepackage{subfigure}
\usepackage{soul}
\usepackage{stfloats}
\usepackage{enumitem}
\usepackage{arydshln}
\usepackage{soul}
\usepackage{mathtools}
\usepackage{amsmath}
\usepackage{amssymb}
\usepackage{amsthm}
\usepackage{multirow}
\usepackage[capitalize,noabbrev]{cleveref}

\allowdisplaybreaks
\theoremstyle{plain}
\newtheorem{theorem}{Theorem}[section]
\newtheorem{proposition}[theorem]{Lemma}
\newtheorem{lemma}[theorem]{Lemma}

\theoremstyle{definition}
\newtheorem{definition}[theorem]{Definition}

\theoremstyle{remark}

\title{Resolving Oversmoothing with Opinion Dissensus}
\author{%
  Keqin Wang$^{1}$ \quad Yulong Yang$^{1}$ \quad Ishan Saha$^{2}$ \quad Christine Allen-Blanchette$^{1,3}$\\
  $^{1}$Mechanical and Aerospace Engineering \quad $^{2}$Electrical and Computer Engineering\\
  $^{3}$Center for Statistics and Machine Learning\\
  Princeton University\\
  \texttt{\{keqin.wang, yulong.yang, ishansaha, ca15\}@princeton.edu}
}
\begin{document}
\maketitle
%%%%%%%%%%%%%%%%%%%%%%%%%%%%%%%%%%%%%%%%%%%%%%%%%%%%%%%
\vspace{-2mm}
\begin{abstract}
\input{sections/0_abstract}
\end{abstract}
%%%%%%%%%%%%%%%%%%%%%%%%%%%%%%%%%%%%%%%%%%%%%%%%%%%%%%%
\section{Introduction}
\label{introduction}
\input{sections/1_introduction}
\section{Related Work}
\input{sections/2_related_work}
%%%%%%%%%%%%%%%%%%%%%%%%%%%%%%%%%%%%%%%%%%%%%%%%%%%%%%%
%\section{Preliminaries}
%\label{Preliminaries}
%\input{sections/3_preliminaries}
%%%%%%%%%%%%%%%%%%%%%%%%%%%%%%%%%%%%%%%%%%%%%%%%%%%%%%%
\section{Graph Neural Networks and Opinion Dynamics}
\label{sec:analogy}
\input{sections/4_understanding_oversmoothing}

%%%%%%%%%%%%%%%%%%%%%%%%%%%%%%%%%%%%%%%%%%%%%%%%%%%%%%%
\section{Behavior-Inspired Message Passing Neural Network}
\label{section:BIMP}
\input{sections/5_BIMP}
%%%%%%%%%%%%%%%%%%%%%%%%%%%%%%%%%%%%%%%%%%%%%%%%%%%%%%%
\section{Empirical Analysis}
\label{section:experiment}
\input{sections/6_experiment}
\section{Conclusion}
\label{section:conclusion}
\input{sections/7_conclusion}
%%%%%%%%%%%%%%%%%%%%%%%%%%%%%%%%%%%%%%%%%%%%%%%%%%%%%%%
\bibliographystyle{unsrt}
\bibliography{main_nips}
%%%%%%%%%%%%%%%%%%%%%%%%%%%%%%%%%%%%%%%%%%%%%%%%%%%%%%%
\appendix
\section{Proofs}
\label{appendix:proof_of_proposition}
\input{appendix/1_proposition_proof}

\section{Datasets}\label{appendix:datasets}
\input{appendix/2_dataset}
\section{Experiment Details}\label{appendix:experiment_details}
\input{appendix/3_experiment_details}
% \section{Additional Experiment}
% \input{appendix/4_additional_experiments}
%%%%%%%%%%%%%%%%%%%%%%%%%%%%%%%%%%%%%%%%%%%%%%%%%%%%%%%
% \input{sections/#_checklist}
%%%%%%%%%%%%%%%%%%%%%%%%%%%%%%%%%%%%%%%%%%%%%%%%%%%%%%%
\end{document}

%% file: sections/0_abstract.tex
While graph neural networks (GNNs) have allowed researchers to successfully apply neural networks to non-Euclidean domains, deep GNNs often exhibit lower predictive performance than their shallow counterparts. This phenomena has been attributed in part to oversmoothing, the tendency of node representations to become increasingly similar with network depth. In this paper we introduce an analogy between oversmoothing in GNNs and consensus (i.e., perfect agreement) in the opinion dynamics literature. We show that the message passing algorithms of several GNN models are equivalent to linear opinion dynamics models which have been shown to converge to consensus for all inputs regardless of the graph structure. This new perspective on oversmoothing motivates the use of nonlinear opinion dynamics as an inductive bias in GNN models. In our Behavior-Inspired Message Passing (BIMP) GNN, we leverage the nonlinear opinion dynamics model which is more general than the linear opinion dynamics model, and can be designed to converge to dissensus for general inputs. Through extensive experiments we show that BIMP resists oversmoothing beyond 100 time steps and consistently outperforms existing architectures even when those architectures are amended with oversmoothing mitigation techniques. We also show that BIMP has several desirable properties including well behaved gradients and adaptability to homophilic and heterophilic datasets.

% In contrast to other classes of neural networks where learned representations become increasingly expressive with network depth, the learned representations in graph neural networks (GNNs) become increasingly similar leading to reduced predictive performance. This phenomena, known as oversmoothing, has been the topic of numerous recent works. In this paper, we propose an analogy between oversmoothing in GNNs and consensus or agreement in opinion dynamics. Through this analogy, we show that the message passing structure of recent continuous-depth GNNs is equivalent to a special case of opinion dynamics (i.e., linear consensus models) which has been theoretically proven to converge to consensus (i.e., oversmoothing) for all inputs. Using the understanding developed through this analogy, we design a new continuous-depth GNN model based on nonlinear opinion dynamics and prove that our model, which we call behavior-inspired message passing neural network (BIMP) circumvents oversmoothing for general inputs. Through extensive experiments, we show that BIMP is robust to oversmoothing and adversarial attack, and consistently outperforms competitive baselines.

%% file: sections/1_introduction.tex
The underlying structure of a broad class of real-world systems is naturally
captured by graphs. For example, in molecules, atoms can be represented by graph nodes and atomic bonds can be represented by edges~\cite{fang2022geometry}; in animal groups, individuals can be represented as nodes and their relative proximity can be represented by edges~\cite{young2013starling}; and in transportation networks, bus stops can be represented as nodes and public transit routes can be represented by edges~\cite{madamori2021latency}.
Because of the broad applicability of graphs, their classification, regression, and generation are of strong interest across scientific communities. While MLPs can be adapted to operate on graph data, graph neural networks (GNNs) are specifically designed to respect the graph permutation symmetry (i.e., invariance to node relabeling) and can therefore learn generalizable graph representations, whereas MLPs are not. 

GNNs have been successfully applied to tasks as diverse as  recommendation~\cite{ying2018graph}, molecular prediction~\cite{wang2022molecular}, protein design~\cite{jha2022prediction,reau2023deeprank}, and computer chip design~\cite{mirhoseini2021graph}. However, GNN representations often become increasingly similar with network depth; a phenomena known as oversmoothing~\cite{li2018deeper, oono2019graph, nt2019revisiting}. 
A number of approaches have been proposed to address oversmoothing in GNNs, including the use of residual connections~\cite{li2018deeper,chen2020simple,liu2020towards,fu2022p}, feature normalization~\cite{zhao2019pairnorm,zhou2020towards,zhou2021understanding}, and alternative architectures~\cite{chiang2019cluster, abu2019mixhop, zeng2019graphsaint}. While many alternative architectures work to incorporate higher-order features\cite{chien2020adaptive, chamberlain2021beltrami,liu2024clifford, li2024generalized}, continuous-depth GNNs instead interpret conventional GNN architectures as discretizations of a continuous process~\cite{poli2019graph, chamberlain2021grand, eliasof2021pde}. This interpretation allows the application of techniques developed for modeling and analyzing dynamical systems~\cite{thompson2002nonlinear, brunton2016discovering, paredes2024output, richards2024output}. % With this insight, a number of authors have turned to physical processes to inspire the design of continuous-depth GNNs to obtain desirable physical properties.
For example, GraphCON~\cite{rusch2022graph} avoids oversmoothing by enforcing stability conditions on hidden states of a coupled and damped oscillator system; and ACMP~\cite{wang2022acmp} avoids oversmoothing by introducing repulsive forces traditionally used to control clustering in particle systems. These are just a few among many that have shown incorporating inductive biases from physical processes can mitigate oversmoothing while maintaining expressivity~\cite{han2023continuous}.

In this paper, we propose a continuous-depth GNN inspired by behavioral interaction, instead of physical processes. First, we introduce an analogy likening node features in a GNN to opinions in an opinion dynamics model, feature aggregation to opinion exchange, and graph outputs to opinion outcomes (Section \ref{sec:analogy}). Using this analogy, we show that oversmoothing will always occur in GNN models with layer-wise aggregation schemes that are equivalent to linear opinion dynamics (Section \ref{sec:understanding_oversmoothing}). With this new understanding of oversmoothing, we leverage the nonlinear opinion dynamics model~\cite{leonard2024fast, bizyaeva2022nonlinear} to design a novel continuous-depth GNN that is provably robust to oversmoothing with desirable characteristics such as well behaved gradients and adaptability to heterophilic datasets (Section \ref{section:BIMP}). Finally, we empirically validate our Behavior-Inspired Message Passing (BIMP) GNN on several datasets and against competitive baselines (Section~\ref{section:experiment}).

%% file: sections/2_related_work.tex
% \paragraph{Oversmoothing in graph neural networks.} 
% Oversmoothing problem is widely noticed and discussed in \cite{li2018deeper, oono2019graph, nt2019revisiting}. As the learned embeddings get similar, oversmoothing can be interpreted as an expressivity decay which is often analyzed using the Weisfeiler-Lehman test \cite{huang2021short}. Recent works provide precise definitions of oversmoothing through various perspectives, including the feature distance \cite{chen2020measuring}, Dirichlet energy \cite{cai2020note,zhou2021dirichlet, rusch2022graph, rusch2023survey} and information gain \cite{zhou2020towards}. 
% Given these definitions, however, finding an efficient method to mitigate oversmoothing remains challenging and attracts significant attention in GNNs research.
% % One intuitive explanation for oversmoothing is node representations get updated by taking a weighted average of its neighbors’ features, which makes nodes perceive similar information across layers.

\textbf{Oversmoothing in GNNs.}
Contrary to conventional feed forward networks~\cite{montufar2014number, lecun2015deep}, deep discrete GNNs suffers degradation of performance from oversmoothing of node features~\cite{li2018deeper, oono2019graph, nt2019revisiting}. Further analysis across wide classes of GNN models have resulted in ways to understand this phenomenon. Focusing on specific architectures, works have revealed that simple linear GNNs exhibits oversmoothing as the added layers exacerbate the denoising and mixing effect of the network~\cite{wu2022non}; GCNs~\cite{kipf2016semi} learns representations that attempts to counteract an inherently oversmoothing prone network structure~\cite{yang2020revisiting}; and attention based networks such as GAT~\cite{velivckovic2017graph} oversmooths at an exponential rate due to the ergodicity of infinite matrix products~\cite{wu2023demystifying}. More generally, oversmoothing have been characterized by energy minimization of gradient flows~\cite{di2022understanding}, representational rank collapse~\cite{roth2024preventing}, and exceeding a theoretical limit of smoothing in mean aggregation~\cite{keriven2022not}.

% Oversmoothing problem is firstly noticed and discussed in \cite{li2018deeper, oono2019graph, nt2019revisiting} and then well studied across GNN model classes. For GCNs, \cite{yang2020revisiting} proposed that GCNs inherently lead to oversmoothing, and the training process actually learns to counteract this. For attention-based GNNs (e.g., GAT), \cite{wu2023demystifying} derived results on the ergodicity of infinite matrix products, showing oversmoothing happens at an exponential rate. For simple linear GNNs, \cite{wu2022non} shows that adding layers strengthens the denoising effect while also exacerbating the mixing effect which leads to oversmoothing. Similarly, \cite{keriven2022not} proved that in the case of mean aggregation, some smoothing is useful but too much leads to oversmoothing. \cite{roth2024preventing} argued that rank collapse is the root cause of oversmoothing, and proposed the sum of Kronecker products (SKP) as a general property that models should exhibit. From a broader perspective, \cite{di2022understanding} interpreted MPNNs with symmetric weights as gradient flows that minimize energy, naturally leading to oversmoothing.

\textbf{Continuous-depth GNNs.} 
Continuous-depth networks such as NeuralODE~\cite{chen2018neural} define the network depth implicitly through the simulation of differential equations. GDE~\cite{poli2019graph} leverages this notion to construct continuous-depth GNNs by propagating inputs through continuum of GNN layers governed by an underlying ODE. In order to better control and understand node dynamics, further work focuses on leveraging physical inductive biases such as heat diffusion~\cite{chamberlain2021grand}, Beltrami flows~\cite{chamberlain2021beltrami}, wave equations~\cite{eliasof2021pde}, coupled damped oscillators~\cite{rusch2022graph}, energy source terms~\cite{thorpe2022grand++}, Allen-Cahn reaction diffusion processes~\cite{wang2022acmp}, blurring-sharpening forces~\cite{choi2023gread}, oscillator synchronization~\cite{nguyen2024coupled}, and Ricci flow~\cite{chen2025graph}. These embedded dynamics provide a systemic approach to counteract known limitations of discrete GNNs such as oversmoothing of deep networks and degradation of performance on heterophilic graphs~\cite{han2023continuous}.

\textbf{Opinion dynamics.} Since opinions can be interpreted as latent preferences, modeling opinion dynamics provide insights into understanding and predicting agents' behavior. Specifically, consensus dynamics~\cite{bullo2018lectures} is commonly used in multi-agent settings such as coordinating multi-vehicle movements~\cite{justh2005natural, leonard2010coordinated}, understanding network systems~\cite{leonard2007collective, ballerini2008interaction}, and learning on graphs~\cite{zhouodnet}. However, the naive implementation of linear models for opinion formulation results in dynamics that only converge to consensus~\cite{altafini2012consensus, dandekar2013biased}. This noted short comings of consensus dynamics models are resolved in nonlinear opinion dynamics~\cite{leonard2024fast}. The nonlinearity results in bifurcations~\cite{golubitsky2012singularities} and hence opinions can evolve to dissensus rapidly even under weak input signals~\cite{bizyaeva2022nonlinear}. Nonlinear opinion dynamics have been shown to able to model a variety of systems such as group decision making~\cite{leonard2021nonlinear, bizyaeva2024active, arango2024opinion}; multi-agent control~\cite{leonard2010coordinated, montes2010opinion}; and relational inference~\cite{yang2024relational}. Our proposed network seeks to integrate nonlinear opinion dynamics as the behavior-inspired inductive bias in a continuous-depth GNN.

%% file: sections/4_understanding_oversmoothing.tex
In this section, we introduce our GNN-opinion dynamics (GNN-OD) analogy. We begin with a review GNNs and opinion dynamics, followed by a brief discussion of bifurcation theory in the context of the nonlinear opinion dynamics model~\cite{leonard2024fast}. We then develop an analogy likening GNNs to opinion dynamics models, and oversmoothing to opinion consensus.

%, and use our analogy to: (1) prove that recent continuous-depth GNN dynamics will exhibit oversmoothing for all inputs; and (2) design a novel continuous-depth GNN, behavior-inspired message passing (BIMP) GNN, using a nonlinear opinion dynamics inductive bias to circumvent oversmoothing and improve expressivity.

\subsection{Graph Neural Networks and Oversmoothing}
\label{sec:GNN}
\textbf{Graph neural networks.} Let $\mathcal{G}=(\mathcal{V},\mathcal{E})$ be a graph with $n=|\mathcal{V}|$ nodes and $m=|\mathcal{E}|$ edges, where an edge $e_{ij}$ exists in $\mathcal{E}$ if the nodes $\mathbf{x}_i$ and $\mathbf{x}_j$ are connected in $\mathcal{G}$. Given an input graph, a graph neural network (GNN) $f:\mathcal{G}\rightarrow\mathcal{Y}$ returns a label (or label set) over edges, nodes, or the entire graph. Of the existing GNN algorithms, a large subset can be described in the message passing framework~\cite{gilmer2017neural}. In this framework, layer-wise transformations are determined by learned message and update functions. The message function at layer $l$, $M^l$, and update function at layer $l$, $U^l$, are of the form,
\begin{equation}
\mathbf{m}^{l+1}_i = \sum_{\mathbf{x}_j \in \mathcal{N}(\mathbf{x}_i)} M^l(\mathbf{x}^l_i, \mathbf{x}^l_j, e_{ij}) \quad\text{and}\quad
\mathbf{x}^{l+1}_i = U^l(\mathbf{x}^{l}_i, \mathbf{m}^{l+1}_i),
\end{equation}
where $\mathbf{x}^l_i$ denotes the representation of node $i$ at layer $l$. %This framework can be used to describe both discrete-time and continuous-time graph neural network models. % Typically, discrete-time MPNNs such as GAT, GCN and xxx, use layer dependent message and update functions are layer dependent, and continuous-time MPNNs, the message and update functions are layer independent.

%\paragraph{Oversmoothing.} 
\textbf{Oversmoothing.} In the GNN literature, oversmoothing is defined as the tendency for node features to become increasingly similar with increasing network depth~\cite{rusch2023survey}. This phenomena has been observed in discrete \cite{wu2022non, yang2020revisiting, wu2023demystifying, keriven2022not} and continuous-depth \cite{chamberlain2021grand, eliasof2021pde,xhonneux2020continuous} GNNs, and correlates with reduced predictive performance. We can measure the degree of oversmoothing in a GNN using the Dirichlet energy~\cite{rusch2023survey, cai2020note}. The Dirichlet energy at layer $l$ is given by,
\begin{equation}
\label{eq:dirichlet_energy}
    E(\mathbf{X}^l)= \frac{1}{n} \sum_{\mathbf{x}_i\in\mathcal{V}} \sum_{\mathbf{x}_j\in\mathcal{N}(\mathbf{x}_i)} \| \mathbf{x}^l_i - \mathbf{x}^l_j\|^2_2,
\end{equation}
where $\mathbf{X}^l=[\mathbf{x}_{1}^{l}, \cdots, \mathbf{x}_{n}^{l}]^{T}$. If the Dirichlet energy tends to zero as $l$ tends to infinity, that is,
\begin{equation}
\label{eq:over-smoothing}
    \lim_{l \to \infty}\|\mathbf{x}_i^l - \mathbf{x}_j^l\|^2_2 = 0 \;\text{for all}\; e_{ij} \in \mathcal{E},
\end{equation} 
the network is said to exhibit oversmoothing.
\subsection{Opinion Dynamics and Opinion Consensus}\label{sec:OD}
%\paragraph{Linear opinion dynamics.}
Let $\mathcal{M}=(\mathcal{G}^a, \mathcal{G}^o)$ be a multi-agent system with a 
communication graph $\mathcal{G}^\mathrm{a} = (\mathcal{V}^\mathrm{a},\mathcal{E}^\mathrm{a},\mathbf{A}^\mathrm{a})$, and an option graph $\mathcal{G}^\mathrm{o} = \left(\mathcal{V}^\mathrm{o}, \mathcal{E}^\mathrm{o}, \mathbf{A}^\mathrm{o}\right)$. 
In this system, each of the $N_\mathrm{a}$ agents has a real-valued opinion on each of the $N_\mathrm{o}$ options.
The adjacency matrix of the communication graph, $\mathbf{A}^{\mathrm{a}}=\left[a^{\mathrm{a}}_{ik}\right]\in\mathbb{R}^{N_{\mathrm{a}}\times N_{\mathrm{a}}}$, defines the communication strength between agents, and the adjacency matrix of the option graph, $\mathbf{A}^\mathrm{o}=[a^{\mathrm{o}}_{jl}]\in\mathbb{R}^{N_{\mathrm{o}}\times N_{\mathrm{o}}}$, defines how correlated different options are. An opinion dynamics model on $\mathcal{M}=(\mathcal{G}^a, \mathcal{G}^o)$ describes the evolution of agent opinions in time.

\textbf{Opinion consensus.} Given a multi-agent system $\mathcal{M}=(\mathcal{G}^a, \mathcal{G}^o)$, a question of interest is whether the opinions of agents tend toward consensus (i.e., perfect agreement). The opinions of agents is said to reach consensus if and only if the opinions of all agents tend toward the same value as time tends to infinity~\cite{degroot1974reaching}, that is,
\begin{equation}
\label{eq:consensus}
    \lim_{t \to \infty}\|\mathbf{x}_i(t) - \mathbf{x}_j(t)\|^2_2 = 0, \;\text{for all}\; e_{ij} \in \mathcal{E}.
\end{equation}
\textbf{Linear opinion dynamics.} The linear opinion dynamics model introduced in~\cite{degroot1974reaching} describes the discrete-time evolution of agent opinions by,
\begin{equation}
\label{eq:lod}
     \mathbf{x}_i(t+1) = \sum_{k=1}^{N_\mathrm{a}} a^{\mathrm{a}}_{ik}\mathbf{x}_k(t),\quad \sum_{k=1}^{N_\mathrm{a}} a^{\mathrm{a}}_{ik} = 1,
\end{equation}
where $a^{\mathrm{a}}_{ik} \geq 0$ can be interpreted as the influence of agent $x_i$ on agent $x_k$, and the total influence of any agent sums to one. In this model, the option graph can be understood as uncorrelated, $a^{\mathrm{o}}_{jl} = 0$. The continuous-time analogue of Equation \eqref{eq:lod} is given by,
\begin{equation}
\label{eq:FD_model}
    \dot{\mathbf{x}}_i(t) = - d_i \mathbf{x}_i(t) + \sum_{k=1}^{N_a} a^{\mathrm{a}}_{ik}\mathbf{x}_k(t),
\end{equation}
where the total influence of agent $i$ is equal to $d_i$~\cite{leonard2024fast}.

In a linear opinion dynamics model, consensus is reached for all initial conditions, and the consensus value is independent of the graph structure and linearly dependent on the initial conditions~\cite{leonard2024fast}. The more general case of linear opinion dynamics with time-varying influence (i.e., $a^{\mathrm{a}}_{ik}(t)$ is time-dependent) can also be shown to converge to consensus~\cite{1393134,7478088, fax2004information, blondel2005convergence}.
\begin{figure}[t]
    \centering
    \includegraphics[width=0.72\linewidth]{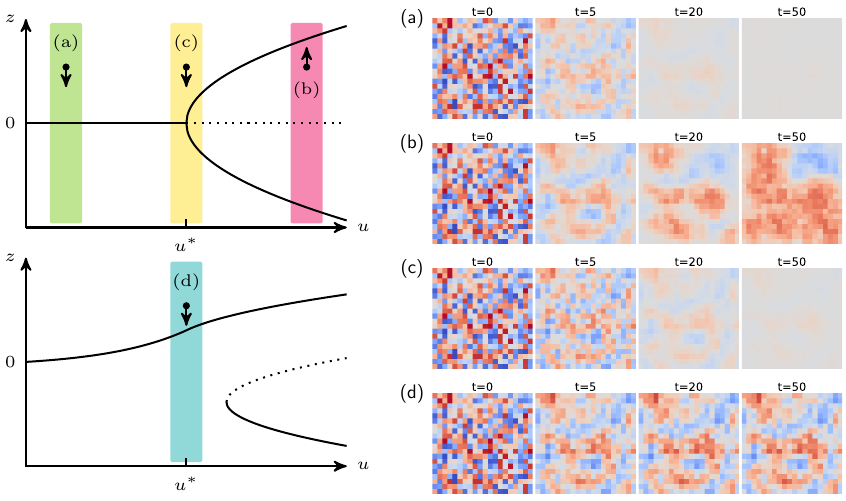}
    \vskip -0.05in
    \caption{\textbf{Nonlinear opinion dynamics and dissensus.} 
    % \textbf{Left:} The bifurcation diagram with $b_{ij}=0$, where the equilibria as a function of attention $u$ (x-axis) and the mapped opinion $z=\langle \mathbf{v,x}\rangle$ (y-axis), with $\mathbf{v}$ denoting the dominant eigenvector of $\mathcal{G}^{\mathrm{a}}$. New equilibria emerge when attention $u$ exceeds the bifurcation point $u^*$. Markers indicate initial opinions used in (a)–(c). \textbf{Right:} Opinion dynamics in a multi-agent system on a grid graph $\mathcal{G}^{\mathrm{a}}$ with one-dimensional opinion ($\mathcal{G}^{\mathrm{o}}=\mathbf{0}$) and random initial opinions. (a) When $u<u^*$, all agents converge to a neutral consensus regardless of initial conditions. (b) With the same initial opinions but with $u > u^*$, the system converges to a dissensus state. (c) Reversing the initial opinions while keeping $u > u^*$ shows the symmetry of the bifurcation. (d) Using the same initial opinion and $u$ but introducing an external force $b_{ij} = x_{ij}(0)$ increases the variance among opinions and converge faster.
    \textbf{(Left)} The pitchfork bifurcation diagram illustrates a change in the number and stability of opinion states with the attention parameter $u$ (stable equilibria are illustrated with a solid line and unstable equilibria are illustrated with a dotted line). In the diagram, $z$ represents the weighted average of agent opinions, and $u^*$ represents the bifurcation point. When the input term $b=0$ we have the pitchfork bifurcation (top-left), and when $b>0$ we have its unfolding (bottom-left). 
    %
    % Points (a)-(d) denote the same initial weighted average opinion under different values of $u$ and $b$. 
    \textbf{(Right)} The time evolution of agent opinions under the nonlinear opinion dynamics model depends on the initial weighted average of agent opinions $z$, the attention parameter $u$, and the weighted average of agent inputs $b$. Each subfigure corresponds to an initial condition on the left. In all cases, the initial $z$ is the same. \textbf{(a)} ($u<u^*$, $b=0$). Agent opinions converge to a neutral consensus (i.e., perfect agreement) which is equivalent to oversmoothing. \textbf{(b)} ($u>u^{*}$, $b=0$). Agent opinions converge to dissensus with low variance, $z$ is positive. \textbf{(c)} ($u=u^{*}$, $b=0$). Agent opinions converge to a neutral consensus. \textbf{(d)} ($u=u^{*}$, $b>0$). Agent opinions converge to dissensus with high variance, $z$ is positive.
    } 
    \label{fig:bifurcation}
    \vskip -0.1in
\end{figure}

\textbf{Nonlinear opinion dynamics.}
The nonlinear opinion dynamics model introduced in \cite{leonard2024fast, bizyaeva2022nonlinear, 10341745} describes the continuous-time evolution of agent $i$'s opinion about option $j$ by,
\begin{equation}
\label{eq:nod}
    \dot{x}_{ij} = -d_{ij}x_{ij} + S\Bigg(u_i \Bigg( \alpha_{ij} x_{ij} + \sum^{N_\mathrm{a}}_{\substack{k=1 \\ k \neq i}}a^{\mathrm{a}}_{ik}x_{kj} + \sum^{N_\mathrm{o}}_{\substack{l=1\\ l\neq j}}a^{\mathrm{o}}_{jl}x_{il} + \sum^{N_\mathrm{a}}_{\substack{k=1 \\ k \neq i}} \sum^{N_\mathrm{o}}_{\substack{l=1 \\ l \neq j}} a^{\mathrm{a}}_{ik} a^{\mathrm{o}}_{jl} x_{kj}\Bigg) \Bigg) + b_{ij},
\end{equation}
where $d_{ij}\geq0$, $u_{i}>0$, and $\alpha_{ij}\geq0$ are parameters intrinsic to the agent, and $a^{\mathrm{a}}_{jk}$, $a^{\mathrm{o}}_{jl}$, and $b_{ij}$ are parameters extrinsic to the agent. 
The intrinsic parameter $d_{ij}$, the damping parameter, describes how resistant agent $i$ is to forming an opinion about option $j$; $u_{i}$, the attention parameter, represents the attentiveness of agent $i$ to the opinions of other agents; and $\alpha_{ij}$, the self-reinforcement parameter, defines how confident agent $i$ is in its opinion on option $j$. The extrinsic parameter $b_{ij}$, the input parameter, represents the impact of the environment on the agent $i$'s opinion about option $j$, and the saturating function $\mathcal{S}$ is selected so that  $S\left(0\right)=0$, $S'\left(0\right)=1$, and $S'''\left(0\right)\neq0$.

By modeling opinion formation as a nonlinear process, the nonlinear opinion dynamics model can capture opinion consensus and dissensus. The nonlinearity induces a bifurcation where the number and/or stability of equilibrium solutions changes (see Figure~\ref{fig:bifurcation} top-left). Consensus results when all agents select exactly the same equilibrium value, dissensus results otherwise. A switch from consensus to dissensus can result from a change in the attention parameter $u$, this can be seen by comparing the dynamics at point (a) and (b) in Figure~\ref{fig:bifurcation} (left), or from a change in the input parameter $b$, this can be seen by comparing the dynamics at point (c) and (d) in Figure~\ref{fig:bifurcation} (left).

\begin{table}[t]
    \caption{\textbf{GNN-OD Analogy.} We describe our analogy between GNNs and opinion dynamics. Our analogy relates the graph structures in GNNs and opinion dynamics, and the notions of oversmoothing and opinion consensus.} 
    \label{tab:consensus_vs_oversmoothing}
    \vskip -0.05in
    \begin{center} 
        \begin{tabular}{ll}
        \toprule
        Graph neural networks & Opinion dynamics \\
        \midrule
        Node & Agent\\
        Edge & Communication link\\
        Node feature dimension & Number of agent options\\
        Value of node $i$ feature $j$ & Opinion of agent $i$ option $j$\\ 
        Oversmoothing & Opinion consensus\\
        \bottomrule
        \end{tabular}
    \end{center}
    \vskip -0.1in
\end{table}

\subsection{Graph Neural Network-Opinion Dynamics Analogy} 
As described in Section \ref{sec:GNN} and \ref{sec:OD}, GNNs and opinion dynamics models % with an unconnected option graph (i.e., $\mathcal{G}^o=\mathbf{0}$) 
have several features which can be understood analogously. The nodes in a GNNs are analogous to the agents in an opinion dynamics model, the edges between nodes are analogous to communication links between agents, and layer-wise oversmoothing is analogous to opinion consensus in time (see Equations \eqref{eq:over-smoothing} and \eqref{eq:consensus}). We summarize our GNN-opinion dynamics (GNN-OD) analogy in Table \ref{tab:consensus_vs_oversmoothing}.

\section{Oversmoothing and Opinion Consensus}
\label{sec:understanding_oversmoothing}
In this section, we use our analogy to prove oversmoothing in several classes of GNNs. We begin with linear discrete-depth GNNs, then continue to Laplacian-based continuous-depth GNNs. The utility of the GNN-OD analogy for understanding oversmoothing, motivates its use in the design of new architectures (Section \ref{section:BIMP}). All proofs are provided in Appendix \ref{appendix:proof_of_proposition}. %, then use our analogy to prove oversmoothing in several GRAND variants \cite{chamberlain2021grand, thorpe2022grand++, rusch2022graph} and {\color{blue} linear mean aggregation GNNs \cite{keriven2022not,kortvelesy2023generalised}.}

\subsection{Linear discrete-depth GNNs}
Linear discrete-depth GNNs (e.g., SGC \cite{wu2019simplifying} and DGC \cite{wang2021dissecting}) can be described using layer-wise transformations of the form,
\begin{equation}
\label{eq:linear_mean_aggregation}
    \mathbf{X}^{l+1} = \mathbf{AX}^{l}\mathbf{W}^{l} =\mathbf{D}\tilde{\mathbf{A}}\mathbf{X}^{l}\mathbf{W}^{l}= \tilde{\mathbf{A}}\mathbf{X}^{l}\mathbf{D}\mathbf{W}^{l},
\end{equation}
where adjacency matrix $\mathbf{A} = \mathbf{D}\tilde{\mathbf{A}}$ for some right stochastic matrix $\tilde{\mathbf{A}}$ and diagonal matrix $\mathbf{D}$ with $\mathbf{D}_{ii} = \sum_{j}\mathbf{A}_{ij}$; and the transformation matrix $\mathbf{W}^l$ is layer dependent. We can use our GNN-OD analogy to show this class of GNNs will exhibit oversmoothing for all inputs, and all input graphs. 

\begin{tcolorbox}[blanker, interior engine=spartan, colback=CornflowerBlue!25, boxsep=1mm, borderline west={1.5pt}{0pt}{gray}]
\begin{lemma}[Linear dynamics oversmooth]
\label{lemma:linear_aggregation}
    Any discrete-depth graph neural network with linear aggregation exhibits oversmoothing.
\end{lemma}
\end{tcolorbox}

Previous works that have shown oversmoothing in various subclasses of linear discrete-depth GNNs %\cite{li2018deeper, wu2022non, keriven2022not}, for example 
include \cite{wu2022non} which proves oversmoothing in convolutional GNNs, and \cite{keriven2022not} which proves oversmoothing in discrete-depth GNNs with linear mean aggregation. %; \hl{to the best of our knowledge, we are the first to show oversmoothing for the general class of linear discrete-depth GNNs.}

\subsection{Laplacian-based continuous-depth GNNs}
\textbf{Laplacian dynamics.} For a given graph, the graph Laplacian of is defined as $\mathbf{L = A - D}$ where $\mathbf{A}$ is the graph adjacency matrix and $\mathbf{D}$ is a diagonal matrix with entries $\mathbf{D}_{ii}=\sum_j \mathbf{A}_{ij}$. GNNs with Laplacian dynamics (e.g., GRAND-$\ell$~\cite{chamberlain2021grand} and GraphCON-Tran~\cite{rusch2022graph}), can be described using the dynamical equation,
\begin{equation}
\label{eq:grand-l}
    \dot{\mathbf{X}}(t) = -\mathbf{LX}(t). 
\end{equation}
We can use our GNN-OD analogy to show this class of GNNs will exhibit oversmoothing for all inputs, and all input graphs.

\begin{tcolorbox}[blanker, interior engine=spartan, colback=CornflowerBlue!25, boxsep=1mm, borderline west={1.5pt}{0pt}{gray}]
\begin{lemma}[Laplacian dynamics oversmooth]
\label{lemma:laplacian_dynamics}
Any continuous-depth graph neural network with Laplacian dynamics exhibits oversmoothing.
\end{lemma}
\end{tcolorbox}

Oversmoothing of continuous-depth GNNs with Laplacian dynamics and time-varying adjacency matrix (e.g., GRAND-$n\ell$~\cite{chamberlain2021grand}), can be shown by analogy with linear opinion dynamics models with time-varying influence (see Section \ref{sec:OD}). Oversmoothing in GRAND-$\ell$~\cite{chamberlain2021grand} was previously shown in \cite{di2022understanding,thorpe2022grand++,choi2023gread}.

\textbf{Laplacian dynamics with an external input.}
Continuous-depth GNNs with Laplacian dynamics and an external input (e.g., \cite{wang2022acmp, thorpe2022grand++, choi2023gread, zhao2023graph,eliasof2023adr}), can be described using the dynamical equation,
\begin{equation}
\label{eq:grand++}
    \dot{\mathbf{X}}(t) = -\mathbf{L}\mathbf{X}(t) + \mathbf{B}(t),
\end{equation}
where $\mathbf{B}(t)$ is the external input. While some works~\cite{wang2022acmp, thorpe2022grand++, choi2023gread} design $\mathbf{B}(t)$ to address oversmoothing, using our GNN-OD analogy, we can understand when oversmoothing occurs.

% While the performance of GRAND++-$\ell$ exceeds that of GRAND-$\ell$, the performance boost comes at the expense of the system equilibrium. Moreover, for large timescales, the difference in node features is relatively small resulting in reduced discriminability, and poorer network performance.

\begin{tcolorbox}[blanker, interior engine=spartan, colback=CornflowerBlue!25, boxsep=1mm, borderline west={1.5pt}{0pt}{gray}]
\begin{proposition}[Laplacian dynamics with an external input oversmooth]
\label{lemma:laplacian_reaction}
     Any continuous-depth graph neural network with Laplacian dynamics and an external input will exhibit oversmoothing when the dynamics are linear.
\end{proposition}  
\end{tcolorbox}

% {\color{blue} This result implies that the normalized Dirichlet energy $\frac{E(t)}{||\mathbf{X}(t)||} \to 0$ as $t \to \infty$ and GRAND++-$\ell$ is Low-Frequency-Dominant (LFD), a notion of oversmoothing defined in \cite{di2022understanding}, indicating adding the source term to this linear opinion dynamics won’t mitigate oversmoothing.}

% Oversmoothing happens since $\mathbf{v}_0$ is a indistinct vector. In other words, while distinctions between nodes exist, they gradually become negligible as the time-dependent term dominates the dynamics, ultimately resulting in oversmoothing.
% But adding bias term does slightly mitigate the oversmoothing since it's barely asymptotic oversmoothing and occurs linearly, much slower than exponentially. % Adding a normalization operation after the final state could fix this issue, but it's not pointed out in origin paper.

% We believe the loss of equilibrium is the culprit, because node features grow so large that the differences between them become negligible as time approaches infinity. 
% Moreover, all nodes have same linear growth rate, making this issue difficult to detect through the calculation of Dirichlet Energy. 
% KuramotoGNN \cite{nguyen2024coupled}, a recent presented continuous-depth model which adds the source term as well. 

%% file: sections/5_BIMP.tex
%%%%%%%%%%%%%%%%%%%%%%%%%%%%%%%%%%%%%%%%%%%%%%%%%%%%%%%%
%
In this section, we describe our Behavior-Inspired Message Passing (BIMP) GNN which leverages the nonlinear opinion dynamics model as an inductive bias. The nonlinear opinion dynamics model is more general than the linear opinion dynamics model, and can be designed to converge to dissensus for general inputs. We begin with our model definition, then prove several desirable properties including robustness to oversmoothing, well behaved gradients, and adaptability to homophilic and heterophilic datasets. All proofs are provided in Appendix \ref{appendix:proof_of_proposition}.

%In this section, we introduce BIMP. We show that by incorporating nonlinear opinion dynamics as an inductive bias, BIMP circumvents oversmoothing while retaining expressivity.
%
% \textbf{Notation.} We use $\otimes$ to denote the Kronecker product. We use $\mathbf{x}_{i}$ to denote the $i$-th row of opinion matrix $\mathbf{X}$, i.e., the opinion of node $i$.
%
%%%%%%%%%%%%%%%%%%%%%%%%%%%%%%%%%%%%%%%%%%%%%%%%%%%%%%%
%
\subsection{Model definition}
Let $\mathcal{G}=(\mathcal{V},\mathcal{E})$ be an input graph with $n$ nodes, where each node has a $d_\text{in}$-dimensional feature representation. BIMP applies a learnable encoder $\phi$, decoder $\psi$, and nonlinear opinion dynamics model to produce an output. The encoder is defined $\phi:\mathbb{R}^{d_{\text{in}}} \to \mathbb{R}^{N_o}$, the decoder is defined $\psi: \mathbb{R}^{N_\mathrm{o}} \to \mathbb{R}^{d_\text{class}}$, and our dynamics are defined by the equation,
\begin{equation}
    \dot{\mathbf{X}}(t)
    = -d \mathbf{X}(t)
    + \tanh\left[u \left( \alpha \mathbf{X}(t) + 
    \mathbf{A}^\mathrm{a} \mathbf{X}(t) 
    + \mathbf{X}(t) {\mathbf{A}^\mathrm{o}}^\top +
    \mathbf{A}^\mathrm{a} \mathbf{X}(t) {\mathbf{A}^\mathrm{o}}^\top
    \right) \right] 
    + \mathbf{X}(0).\label{eq:BIMP}
\end{equation}
In our dynamics, the parameters $d$ and $\alpha$ are hyperparameters, the attention parameter $u=\frac{d}{\alpha + 3}$, the initial condition $\mathbf{X}(0)=\phi(\mathbf{X}_\text{in})$, and the adjacency matrices of communication and option graphs, $\mathbf{A}^\mathrm{a}$ and $\mathbf{A}^\mathrm{o}$ respectively, are learned. The output of our model is given by $\mathbf{Y} = \psi(\mathbf{X}(T))$, where the terminal time $T$ is a hyperparameter.

% learnable adjacency matrices based on self-attention mechanism \cite{vaswani2017attention}.
%
% The numerical integration method used to solve the ODE is selected based on the dataset and can be one of forward Euler, Runge-Kutta 4th order (RK4), or the Dormand-Prince (Dopri5) method.
%
% {\color{blue} compared with Equation \ref{eq:nod}}. 
% %
% {\color{blue} We propose that our BIMP model possesses greater expressive capacity than a broad class of continuous-depth GNNs, for two key reasons. First, the flexibility of nonlinear opinion dynamics allows BIMP to model more diverse node feature representations than approaches whose dynamics are linear in nature. Second, the belief matrix $ {\mathbf{A}^\mathrm{o}}$ enables BIMP to incorporate feature mixing mechanism, which usually is absent in a wide range of continuous-depth GNNs.}
%
\iffalse
The flexibility of nonlinear opinion dynamics allows BIMP to model more diverse node feature representations than approaches whose dynamics are linear in nature.
%
In fact, under certain assumptions, GRAND-$\ell$ can be viewed as the first order approximation of BIMP, which intuitively indicates that BIMP possesses greater expressive capacity than GRAND-$\ell$ its derivatives.
%
\begin{proposition}\label{prop:first_order_same_grand}
    For a BIMP model with attention parameter $u=1$, bias parameter $\mathbf{B=0}$, and uncorrelated options $\mathbf{A^\mathrm{o}}=\mathbf{0}$, the first-order approximation is equivalent to GRAND-$\ell$.
\end{proposition}
\fi

\subsubsection{The communication and option graphs.}
The communication and option graphs are designed to allow for theoretical analysis, and reduce computational expense. 
For a nonlinear opinion dynamics model of the form,
\begin{equation}
\label{eq:uncorrelated_options}
    \dot{\mathbf{X}} = -d\mathbf{X} + \tanh\left[ u(\alpha \mathbf{X} + \mathbf{A^\mathrm{a}} \mathbf{X}) \right] + \mathbf{B}, 
\end{equation}
the communication adjacency matrix $\mathbf{A}^\mathrm{a}$ must have a simple leading eigenvalue to admit analysis~\cite{leonard2024fast}.
Our BIMP model described in Equation~\eqref{eq:BIMP} can be written in the form of Equation~\eqref{eq:uncorrelated_options} by combining the communication and option graphs into a single effective adjacency matrix.
% We introduce the effective communication matrix to model influence from both belief and communication graphs.
%
\begin{tcolorbox}[blanker, interior engine=spartan, colback=Thistle!25, boxsep=1mm, borderline west={1.5pt}{0pt}{gray}]
\begin{definition}[Effective adjacency matrix]
\label{def:kroneker} Given the adjacency matrices of the communication and option graphs, $\mathbf{A}^\mathrm{a}$ and $\mathbf{A}^\mathrm{o}$, the effective adjacency matrix $\tilde{\mathbf{A}}$ is defined,
    \begin{equation}\label{eq:tilde_a}
        \tilde{\mathbf{A}}=(\mathbf{A}^\mathrm{o} + \mathbf{I})\otimes (\mathbf{A}^\mathrm{a} + \mathbf{I}).
    \end{equation}
\end{definition}
\end{tcolorbox}
%
% Leveraging the simplification provide in this definition, the general form of nonlinear opinion dynamics given in Equation~\eqref{eq:BINN} can be reduced to the uncorrelated form given in Equation~\eqref{eq:uncorrelated_options}.
%
\begin{tcolorbox}[blanker, interior engine=spartan, colback=CornflowerBlue!25, boxsep=1mm, borderline west={1.5pt}{0pt}{gray}]
\begin{proposition}
\label{prop:uncorrelated_equation}
    The general form of nonlinear opinion dynamics (Equation~\eqref{eq:BIMP}) can be written,
    \begin{equation}\label{eq:BIMP_uncorrelated}
        \dot{\mathbf{x}} = -d\mathbf{x} + \tanh\left[ u\biggl((\alpha - 1)\mathbf{x} + \tilde{\mathbf{A}} \mathbf{x}\biggr) \right] + \mathbf{b},
    \end{equation}
    where $\tilde{\mathbf{A}}$ is the effective adjacency matrix, $\mathbf{x} = \operatorname{vec}(\mathbf{X})$ and $\mathbf{b} = \operatorname{vec}(\mathbf{B})$.
\end{proposition}
\end{tcolorbox}
In order to analyze the behavior of BIMP, the effective adjacency matrix must be constrained to have a simple leading eigenvalue. We enforce this condition by constraining the learned effective adjacency matrix to be right stochastic (the leading eigenvalue of a right stochastic matrix is $\lambda_{\max}=1$). 
Learning this matrix directly would be computationally prohibitive (the size of the effective adjacency matrix is $\mathbb{R}^{N_\mathrm{o}N_\mathrm{a}\times N_\mathrm{o}N_\mathrm{a}}$).
To relieve the computational burden, we instead learn the communication and option graphs separately.
Specifically, the entries of $\mathbf{A}^{\mathrm{a}} = [a^{\mathrm{a}}_{ik}]$ and $\mathbf{A}^{\mathrm{o}} = [a^{\mathrm{o}}_{jl}]$ are defined using multi-head attention,
\begin{equation}
    a^{\mathrm{a}}_{ik} = \operatorname{softmax}\left( \frac{(\mathbf{W}^{\mathrm{a}}_K\mathbf{x}_i)^\top \mathbf{W}^{\mathrm{a}}_Q\mathbf{x}_k}{d_k^\mathrm{a}} \right), \quad
    a^{\mathrm{o}}_{jl} = \operatorname{softmax}\left( \frac{(\mathbf{W}^{\mathrm{o}}_K\mathbf{x}^\top_j)^\top \mathbf{W}^{\mathrm{o}}_Q\mathbf{x}^\top_l}{d_k^\mathrm{o}} \right),\label{eq:attention_a}
\end{equation}
where $\mathbf{W}^{\mathrm{a}}_K$, $\mathbf{W}^{\mathrm{a}}_Q$, $\mathbf{W}^{\mathrm{o}}_K$ and $\mathbf{W}^{\mathrm{o}}_Q$ are the key and query weight matrices for communication and option graphs.
A useful consequence of this approach is that the leading eigenvalue $\lambda^{\tilde{a}}_{\max}$ of $\tilde{\mathbf{A}}$ is constant and does not need to be recomputed during training.

\begin{tcolorbox}[blanker, interior engine=spartan, colback=CornflowerBlue!25, boxsep=1mm, borderline west={1.5pt}{0pt}{gray}]
\begin{proposition}
\label{proposition:lambda}
    The leading eigenvalue of the effective adjacency matrix $\tilde{\mathbf{A}}$ is $\lambda^{\tilde{a}}_{\max}=4$.
\end{proposition}
\end{tcolorbox}
\subsection{Parameter selection}
%In the nonlinear opinion dynamics model, the attention parameter $u$ and the input parameter $b$ 
\textbf{The attention parameter $u$.}\label{sec:u_parameter}
In the nonlinear opinion dynamics model, the attention parameter $u$ is the bifurcation parameter. 
Near the bifurcation point $u^*$ (i.e., the point where the number and/or stability of solutions change), the model is ultrasensitive to the input $\mathbf{B}$, and agents will quickly form an opinion (see Figure~\ref{fig:bifurcation}). We design BIMP to be ultrasensitive to the input by setting the value of the attention parameter $u$ to the bifurcation point of the attention-opinion bifurcation diagram.

% In order to maintain the expressive capacity of BIMP, it must exist on the nonlinear regime of Equation~\eqref{eq:BINN}.
%
% The degeneration to a linear model occurs under two settings:
%
% (1) when $u$ is very small and the nonlinear term evaluates to 0;
%
% or (2) when $u$ is very large such that the hyperbolic tangent saturates, and therefore the nonlinear term evaluates to $\pm1$.
%
% To avoid both these situations, we set the attention $u$ to the critical point of the attention-opinion bifurcation {\color{blue} where new equilibria emerge and equilibrium $\mathbf{X}=\mathbf{0}$ switches from stable to unstable.} (see \changes{Appendix xxx}).
%
\begin{tcolorbox}[blanker, interior engine=spartan, colback=CornflowerBlue!25, boxsep=1mm, borderline west={1.5pt}{0pt}{gray}]
\begin{proposition}[Bifurcation point $u^*$]
\label{proposition:u}
    The bifurcation point $u^*$ of the attention-opinion bifurcation diagram is equal to $\nicefrac{d}{\left(\alpha-1 + \lambda^{\tilde{a}}_{\max}\right)}$.
\end{proposition}
\end{tcolorbox}
From Lemma \ref{proposition:u} and \ref{proposition:lambda}, the value of the attention parameter at the bifurcation point is $u= \frac{d}{\alpha+3}$.

\textbf{The input parameter $\mathbf{B}$.} 
In the nonlinear opinion dynamics model, the input parameter $\mathbf{B}$ transforms the bifurcation diagram from a symmetric pitchfork bifurcation to an unfolded pitchfork bifurcation (see Figure~\ref{fig:bifurcation}). This is a form of selective ultrasensitivity where the directions of ultrasensitivity are determined by the structure of the communication graph \cite{bizyaeva2022nonlinear,leonard2024fast}. 

In BIMP with effective adjacency matrix $\tilde{\mathbf{A}}$ and attention parameter $u=\frac{d}{\alpha+3}$, oversmoothing depends on the choice of input parameter. We select the input parameter $\mathbf{B}$, so that BIMP converges to an equilibrium for all initial opinions. 
% When the bias parameter is constant, BIMP converges to equilibrium.

% With our choice of effective adjacency matrix $\tilde{\mathbf{A}}$ and attention parameter $u$, BIMP converges to equilibrium when the input parameter $\mathbf{B}$ is constant. % {\color{blue} and designed bias $\mathbf{b}$,} result in a BINN model that converges to {\color{blue} consensus ($\mathbf{b=0}$) or} dissensus {\color{blue} ($\mathbf{b\neq 0}$) }. 
% We can calculate the equilibrium and verify if it converge at consensus (oversmoothing).
% A BINN model without bias has oversmoothing. Noticing the oversmoothing always occurs when the system reach the equilibrium (it's obvious since the dynamical system evolves over time, if oversmoothing occurs before equilibrium is reached, then it can be easily broken by keeping evolution), in the following part we would only focus on the equilibrium.

\begin{tcolorbox}[blanker, interior engine=spartan, colback=CornflowerBlue!25, boxsep=1mm, borderline west={1.5pt}{0pt}{gray}]
\begin{proposition}[BIMP converges to equilibrium]
\label{proposition:stability}
    A BIMP model with a constant input parameter $\mathbf{B}$ converges to an equilibrium.
\end{proposition}
\end{tcolorbox}

A consequence of this design choice is that we can now understand the oversmoothing behavior of BIMP by analyzing its equilibrium behavior. When the input parameter $\mathbf{B}$ is nonzero, BIMP will not exhibit oversmoothing. 
% We prove Proposition \ref{proposition:stability} in Appendix \ref{appendix:stability} by showing that the nonlinear opinion dynamics are monotone and bounded. Consequently, there are no dynamical behaviors other than convergence to an equilibrium. 

\begin{tcolorbox}[blanker, interior engine=spartan, colback=CornflowerBlue!25, boxsep=1mm, borderline west={1.5pt}{0pt}{gray}]
\begin{theorem}[Dissensus in BIMP]
    \label{theorem:b!=0}
    BIMP will not exhibit oversmoothing when the input parameter $\mathbf{B}$ is time-independent with unique entries. 
\end{theorem}
\end{tcolorbox}

To ensure $\mathbf{B}$ satisfies the conditions of Theorem \ref{theorem:b!=0}, we set $\mathbf{B} = \mathbf{X}(0)$. A similar strategy is used GRAND++~\cite{thorpe2022grand++}, GREAD~\cite{choi2023gread} and KuramotoGNN~\cite{nguyen2024coupled}. 
%
%%%%%%%%%%%%%%%%%%%%%%%%%%%%%%%%%%%%%%%%%%%%%%%%%%%%%%%
%
\subsection{Emergent properties}

\begin{table}[t]
    \caption{\textbf{Classification accuracy on homophilic datasets.} Classification accuracy on the Cora, citeseer, Pubmed, CoauthorCS, Computers, and Photo datasets are reported. Our BIMP model outperforms competitive baselines on all datasets. We highlight the {\setlength{\fboxsep}{1pt}\colorbox{LimeGreen!50}{\textbf{best}}} and {\setlength{\fboxsep}{1pt}\colorbox{Goldenrod!50}{second best}} accuracy.}
    \label{tab:acc_result_homo}
    \centering
    \setlength\tabcolsep{2.2pt}
        \begin{tabular}{lcccccccc}
        \toprule
        Dataset & Cora & citeseer & Pubmed & CoauthorCS & Computers & Photo \\
        \midrule
        \textbf{BIMP} & {\setlength{\fboxsep}{1pt}\colorbox{LimeGreen!50}{\textbf{83.19$\pm$1.13}}} & {\setlength{\fboxsep}{1pt}\colorbox{LimeGreen!50}{\textbf{71.09$\pm$1.40}}} & {\setlength{\fboxsep}{1pt}\colorbox{LimeGreen!50}{\textbf{80.16$\pm$2.03}}} & {\setlength{\fboxsep}{1pt}\colorbox{LimeGreen!50}{\textbf{92.48$\pm$0.26}}} & {\setlength{\fboxsep}{1pt}\colorbox{LimeGreen!50}{\textbf{84.73$\pm$0.61}}} & {\setlength{\fboxsep}{1pt}\colorbox{LimeGreen!50}{\textbf{92.90$\pm$0.44}}} \\
        GRAND-$\ell$ & 82.20$\pm$1.45 & 69.89$\pm$1.48 & 78.19$\pm$1.88 & 90.23$\pm$0.91 & {\setlength{\fboxsep}{1pt}\colorbox{Goldenrod!50}{82.93$\pm$0.56}} & 91.93$\pm$0.39 \\
        GRAND++-$\ell$ & {\setlength{\fboxsep}{1pt}\colorbox{Goldenrod!50}{82.83$\pm$1.31}} & 70.26$\pm$1.46 & {\setlength{\fboxsep}{1pt}\colorbox{Goldenrod!50}{78.89$\pm$1.96}} & 90.10$\pm$0.78 & 82.79$\pm$0.54 & 91.51$\pm$0.41 \\
        KuramotoGNN & 81.16$\pm$1.61 & {\setlength{\fboxsep}{1pt}\colorbox{Goldenrod!50}{70.40$\pm$1.02}} & 78.69$\pm$1.91 & {\setlength{\fboxsep}{1pt}\colorbox{Goldenrod!50}{91.05$\pm$0.56}} & 80.06$\pm$1.60 & {\setlength{\fboxsep}{1pt}\colorbox{Goldenrod!50}{92.77$\pm$0.42}} \\
        GraphCON-Tran & 82.80$\pm$1.34 & 69.60$\pm$1.16 & 78.85$\pm$1.53 & 90.30$\pm$0.74 & 82.76$\pm$0.58 & 91.78$\pm$0.50 \\
        GAT & 79.76$\pm$1.50 & 67.70$\pm$1.63 & 76.88$\pm$2.08 & 89.51$\pm$0.54 & 81.73$\pm$1.89 & 89.12$\pm$1.60 \\
        GCN & 80.76$\pm$2.04 & 67.54$\pm$1.98 & 77.04$\pm$1.78 & 90.98$\pm$0.42 & 82.02$\pm$1.87 & 90.37$\pm$1.38 \\
        GraphSAGE & 79.37$\pm$1.70 & 67.31$\pm$1.63 & 75.52$\pm$2.19 & 90.62$\pm$0.42 & 76.42$\pm$7.60 & 88.71$\pm$2.68 \\
        \bottomrule
    \end{tabular}
    \vskip -0.2in
\end{table}
\begin{table}[b]
    \vskip -0.1in
    \centering
    \caption{ \textbf{Classification accuracy on heterophilic datasets.} Classification accuracy on the Texas, Wisconsin, and Cornell datasets are reported, where BIMP outperforms competitive baselines.}
    \label{tab:acc_result_hetero}
    \setlength\tabcolsep{2.2pt}
    \begin{tabular}{lccc}
        \toprule
        Dataset &  Cornell &  Texas &  Wisconsin \\
        \textit{Homophily level} & 0.30 & 0.11 & 0.21 \\
        \midrule
        \textbf{BIMP} &  {\setlength{\fboxsep}{1pt}\colorbox{LimeGreen!50}{\textbf{77.13$\pm$3.38}}}  & {\setlength{\fboxsep}{1pt}\colorbox{LimeGreen!50}{\textbf{82.16$\pm$4.06}}} & {\setlength{\fboxsep}{1pt}\colorbox{LimeGreen!50}{\textbf{86.57$\pm$4.33}}} \\
        GRAND-$\ell$ & 70.00$\pm$6.22 &   74.59$\pm$5.43 &  82.75$\pm$3.90\\
        GRAND++-$\ell$ & 70.30$\pm$8.50 & 76.14$\pm$5.77 & 83.09$\pm$2.83 \\
        KuramotoGNN & {\setlength{\fboxsep}{1pt}\colorbox{Goldenrod!50}{76.02 $\pm$2.77}} & {\setlength{\fboxsep}{1pt}\colorbox{Goldenrod!50}{81.81$\pm$4.36}} &  {\setlength{\fboxsep}{1pt}\colorbox{Goldenrod!50}{85.09$\pm$4.42}}\\
        GraphCON-GCN &74.05$\pm$3.24   & 80.54$\pm$4.49& 84.79$\pm$2.51  \\
        GAT & 42.16$\pm$7.07 & 57.84$\pm$5.82 & 49.61$\pm$4.21 \\
        GCN & 41.35$\pm$4.69 & 57.03$\pm$5.98 & 48.43$\pm$5.75 \\
        GraphSAGE & 70.54$\pm$2.55 & 72.70$\pm$5.47 & 73.14$\pm$6.27 \\
        \bottomrule
    \end{tabular}
\end{table}

\textbf{Well behaved gradients.}
The stability of neural network gradients impacts training efficiency and learning outcomes \cite{rusch2022graph, nguyen2024coupled,pascanu2013difficulty,awasthi2021convergence,arroyo2025vanishing}. Exploding gradients make learning unstable and vanishing gradients slow down learning. In GNNs, oversmoothing has been shown to occur when network gradients vanish~\cite{rusch2022graph}. In BIMP, we find that the structure of the nonlinear opinion dynamics inductive bias yields bounded gradients that do not vanish exponentially, even for very deep architectures.

% Gradient exploding and vanishing is a vital topic in learning process, especially when the model's architecture is extremely deep.
% Additionally, \cite{rusch2022graph} emphasize the oversmoothing happens when the graph gradients vanish quickly. 
% Increasing the number of layers also lead to other issues, e.g., gradient exploding or vanishing. 
% Even for very deep networks, BIMP gradients will not explode or vanish exponentially during training. %is designed to mitigate gradient exploding and vanishing. 
% We demonstrate the gradient exploding and exponential gradient vanishing won't happen.
\begin{tcolorbox}[blanker, interior engine=spartan, colback=CornflowerBlue!25, boxsep=1mm, borderline west={1.5pt}{0pt}{gray}]
\begin{theorem}[BIMP has well behaved gradients]\label{theorem:gradient}
BIMP gradients are upper bounded and do not vanish exponentially.
\end{theorem}
\end{tcolorbox}

% Additionally, we can rewrite the gradient descent equation with chain rule into a repeated sum operation, alleviating the gradient exponential vanishing. 

% Therefore, BIMP maintains well behaved gradients even for very deep architectures. % BIMP maintains a reliable trainability even for extremely deep models. 
% We prove Proposition \ref{proposition:gradient_explode} and \ref{proposition:gradient_vanish} in Appendix \ref{appendix:gradient}. 

\textbf{Model adaptability.}
In many GNNs, neighborhood aggregation can be interpreted as low-pass filtering~\cite{nt2019revisiting,bo2021beyond,balcilar2021analyzing}. A direct consequence is that these same GNNs will perform poorly on heterophilic datasets (i.e., datasets where edges in an input graphs connect dissimilar nodes). To address this issue, previous works have incorporated high-pass filters which have a sharpening effect~\cite{han2023continuous,di2022understanding,choi2023gread}. In BIMP, we find that the nonlinear opinion dynamics inductive bias can be interpreted as a tunable filter.  
%
% Many GNNs perform poorly on heterophilic datasets, where edges link dissimilar nodes, due to their inherent low-pass filtering nature in graph signal processing \cite{nt2019revisiting,bo2021beyond,balcilar2021analyzing}. 
% Incorporating high-pass filters, which serves as a sharpen effect, can mitigate this issue \cite{di2022understanding,han2023continuous,choi2023gread}.

This becomes clear by writing the BIMP dynamics, defined in Equation \eqref{eq:BIMP_uncorrelated}, in the alternative form, 
\begin{equation}
\label{eq:BIMP_filter}
    \dot{\mathbf{x}} = -d \mathbf{x} + 
    \tanh\Bigl[ u \bigl( (\alpha - 1) \underbrace{(\mathbf{x - \tilde{A} x})}_{\mathclap{\text{sharpening}}} + \alpha \underbrace{\vphantom{(}\mathbf{\tilde{A} x}}_{\mathclap{\text{smoothing}}} \bigr) \Bigr] + \mathbf{b}.
\end{equation}
In this form, BIMP has a high pass filter when $\alpha > 1$ ($(\mathbf{x - \tilde{A} x)}$ sharpens the features); and only a low-pass filter when $\alpha \leq 1$ ($\mathbf{\tilde{A} x}$ smooths the feature). By tuning the hyperparameter $\alpha$, BIMP can be adapted to both homophilic and heterophilic datasets.

% Additionally, Theorem \ref{theorem:b!=0} shows that the input term $\mathbf{b}$ acts as an anti-smoothing force \cite{wang2022acmp,di2022understanding, choi2023gread,han2022generalized}, pulling features apart to counter smoothing effect. This behavior is illustrated in Figure \ref{fig:bifurcation} (d).

%% file: sections/6_experiment.tex
\begin{figure}[t]
    \centering
    \includegraphics[width=\textwidth]{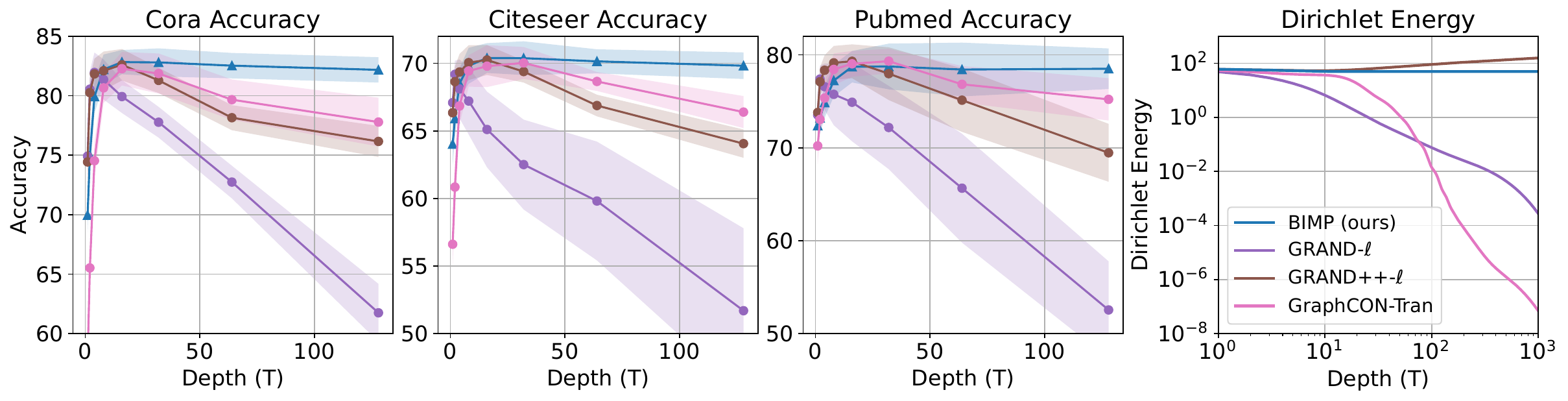}
    \vskip -0.05in
    \caption{\textbf{Classification accuracy and Dirichlet energy.} BIMP is designed to learn node representations that resist oversmoothing even for very large depths. \textbf{(Left)} We compare the classification accuracy of BIMP to baseline models for architectures with $1, 2, 4, 8, 16, 32, 64$ and $128$ timesteps. Our BIMP model is stable out to 128 timesteps, while baseline performance deteriorates after 32 timesteps. \textbf{(Right)} We compare the Dirichlet energy of node features over a range of network depths. The Dirichlet energy of BIMP remains stable even at very deep layers, while the energy of baseline modes does not. }
    \label{fig:depth}
\end{figure}

In this section, we highlight the classification accuracy of BIMP on homophilic and heterophilic datasets; and the robustness of BIMP features to oversmoothing even in very deep architectures. 

\subsection{Classification accuracy}
\label{subsec:classification}
\textbf{Homophilic datasets.} We report the classification performance of our BIMP model and competitive baselines on the Cora~\cite{mccallum2000automating}, citeseer~\cite{sen2008collective}, Pubmed~\cite{namata2012query}, CoauthorCS~\cite{shchur2018pitfalls}, Computers~\cite{shchur2018pitfalls}, and Photo~\cite{shchur2018pitfalls} datasets in Table~\ref{tab:acc_result_homo}. We report the performance of baselines amended with various oversmoothing mitigation techniques in Appendix~\ref{homophilic_appendix}. Our BIMP model outperforms all baseline and amended baseline models on all datasets.

\textbf{Heterophilic datasets.} We report the classification performance of our BIMP model and competitive baselines on the Texas, Wisconsin, and Cornell datasets from the CMU WebKB~\cite{craven1998learning} project in Table \ref{tab:acc_result_hetero}. We report additional comparisons against baselines amended with various oversmoothing mitigation techniques in Appendix~\ref{heterophilic_appendix}. Our BIMP model outperforms all baseline and amended baseline models on all datasets.

\textbf{Large graphs.} We report the classification performance of BIMP and competitive baselines on the ogbn-arXiv~\cite{hu2020open} large graph dataset in Appendix \ref{appendix:large_graph}, Table \ref{tab:large_graph}. BIMP consistently outperforms all baseline models.

\subsection{Performance at large depths}
\label{experiment_depth}
\textbf{Classification accuracy.} To understand the impact of depth on classification accuracy, we compare the performance of our BIMP model to baseline models over a range of network depths (hyperparameters remain fixed across depths). We report classification performance in Figure \ref{fig:depth} (left). BIMP performs comparably to baseline models for shallow depths ($l<32$) and consistently outperforms all baseline models at greater depths.
%, GRAND-$\ell$ , GRAND++-$\ell$ and GraphCON-Tran on the Cora~\cite{mccallum2000automating}, citeseer~\cite{sen2008collective} and Pubmed~\cite{namata2012query} datasets. Each model is trained over a range of network depths and fixed hyperparameters; results are reported in Figure \ref{fig:depth} left. Our BIMP model has comparable performance with baseline models when networks depths are small, but increasingly better performance compared to baseline models as network depths increase. 
Experimental details are provided in Appendix \ref{appendix:experiment_depth}. We report additional comparison against discrete-depth GNNs and baselines amended with various oversmoothing mitigation techniques in Appendix~\ref{appendix:experiment_depth}.

\textbf{Dirichlet energy.} To understand how oversmoothing evolves with depth, we compare the Dirichlet energy (see Equation~\eqref{eq:dirichlet_energy}) of our BIMP model to baseline models over 1000 timesteps and plot the results in Figure \ref{fig:depth} (right). The Dirichlet energy of our BIMP model is stable for all timesteps while the energy of baseline models diverges. Additional experimental details are provided in Appendix \ref{appendix:experiment_dirichlet}.
We report additional comparison against discrete-depth GNNs and baselines amended with various oversmoothing mitigation techniques in Appendix~\ref{appendix:experiment_dirichlet}.

%% file: sections/7_conclusion.tex
In this paper, we propose an analogy between GNNs and opinion dynamics models highlighting the equivalence between oversmoothing in GNNs and opinion consensus in opinion dynamics. Through our analogy, we prove that several existing GNN algorithms are equivalent to linear opinion dynamics models which converges to consensus for all initial conditions and graph structures. Motivated by this, we introduce a novel class of continuous-depth GNNs called Behavior-inspired message passing (BIMP) GNNs which leverage the nonlinear opinion dynamics inductive bias which improves expressivity and guarantees dissensus. 
Experiments against competitive baselines illustrate our model's strong performance and robustness to oversmoothing.

\textbf{Limitations.} The nonlinear opinion dynamics inductive bias may introduce training instabilities at larger step sizes. For example, when using the Euler method, if the step size is larger than $1/d$, where $d$ is the damping term in our dynamical update equation (see Equation~\eqref{eq:BIMP}), each update will lead to a sign change. Since the $d$ is a hyperparameter this is not a severe limitation, and should be navigable in most settings.

\newpage

% dynamical equation is evaluated using a numerical integrator. integrator operates 
% With large step sizes, the damping term $d$ leads to training instability, limiting training acceleration with large time scales.

% , extremely large step-size would make the learning unstable and even fail, limiting the acceleration of training. 
% Also, in the Experiment \ref{experiment:attack}, we observe the promising potential of BIMP in resisting adversarial attack. 
% Future work will develop deeper insights into the effect of the nonlinearity in BIMP on its resilience to adversarial attacks.

%% file: appendix/1_proposition_proof.tex
\subsection{Oversmoothing and Opinion Consensus}
\subsubsection{Lemma \ref{lemma:linear_aggregation}: Linear dynamics oversmooth}

\begin{tcolorbox}[blanker, interior engine=spartan, colback=CornflowerBlue!25, boxsep=1.5mm, borderline west={1.5pt}{0pt}{gray}]
\textit{Any discrete-depth graph neural network with linear aggregation exhibits oversmoothing.}
\end{tcolorbox}
\begin{proof}
Without loss of generality, we consider the case where $\mathbf{A}$ is right stochastic. In this case, we can write the output of the $L$-th layer,
\begin{equation}
    \mathbf{X}^{L} = \mathbf{AX}^{L-1}\mathbf{W}^{L}= \mathbf{A}^{(L)}\mathbf{X}^{0}\mathbf{W}^0\cdots\mathbf{W}^L.
\end{equation}
where $\mathbf{A}^{(n)}$ denotes the $n$-th power of matrix $\mathbf{A}$.

By our GNN-OD analogy, the quantity $\mathbf{A}^{(L)}\mathbf{X}^{0}$ tends toward consensus for all initial conditions. Since this cannot be changed by a linear transformation (e.g., the product $\mathbf{W}^0\cdots\mathbf{W}^L$),  $\mathbf{X}^{L}$ will also tend toward consensus. Since oversmoothing is analogous to opinion consensus, any discrete-time GNN with linear aggregation will exhibit oversmoothing for all inputs.
\end{proof}

\subsubsection{Lemma \ref{lemma:laplacian_dynamics}: Laplacian dynamics oversmooth}
\begin{tcolorbox}[blanker, interior engine=spartan, colback=CornflowerBlue!25, boxsep=1.5mm, borderline west={1.5pt}{0pt}{gray}]
\textit{Any continuous-depth graph neural network with Laplacian dynamics exhibits oversmoothing.}
\end{tcolorbox}

\begin{proof}
    Any GNN with Laplacian dynamics can be expressed in the form,
    \begin{equation}
        \dot{\mathbf{X}}(t) = -\mathbf{DX}(t) + \mathbf{AX}(t).
    \end{equation}
    In this form, it is clear that the dynamics are equivalent to linear opinion dynamics (see Equation \eqref{eq:FD_model}). Since in a linear opinion dynamics model consensus is reached for all initial conditions, a GNN with Laplacian dynamics will also exhibit oversmoothing for all inputs.
\end{proof}

\subsubsection{Lemma \ref{lemma:laplacian_reaction}: Laplacian dynamics with an external input oversmooth} 
\begin{tcolorbox}[blanker, interior engine=spartan, colback=CornflowerBlue!25, boxsep=1.5mm, borderline west={1.5pt}{0pt}{gray}]
\textit{Any continuous-depth graph neural network with Laplacian dynamics and an external input will exhibit oversmoothing when the dynamics are linear. }
\end{tcolorbox}

\begin{proof}
We prove Lemma \ref{lemma:laplacian_reaction} by demonstrating potential oversmoothing behavior in two popular methods: GRAND++$\ell$~\cite{thorpe2022grand++} (in Lemma \ref{lemma:grand++}), GREAD-F and GREAD-FB*~\cite{choi2023gread} (in Lemma \ref{lemma:gread}). The insight that the stability of linear dynamics is sensitive to the external input can be generalized to the analysis of other Laplacian dynamics with an external input.
\end{proof}

\begin{tcolorbox}[blanker, interior engine=spartan, colback=CornflowerBlue!25, boxsep=1mm, borderline west={1.5pt}{0pt}{gray}]
\begin{proposition}[Oversmoothing in GRAND++-$\ell$]\label{lemma:grand++}
    GRAND++-$\ell$ exhibits oversmoothing.
\end{proposition}
\end{tcolorbox}

\begin{proof}
In GRAND++-$\ell$, the layer-wise transformations is of the form,
\begin{equation}\label{eq:grand++_appendix}
    \dot{\mathbf{X}} = -\mathbf{L}\mathbf{X} + \mathbf{B},
\end{equation}
where $\mathbf{B}$ is the fixed source term that depends on the initial state $\mathbf{X}(0)$. 
Defining the right eigenvector matrix $\mathbf{T} = [\mathbf{v}_1, \mathbf{v}_2, ...,\mathbf{v}_{N_a}]$ and left eigenvector matrix $\mathbf{T}^{-1} = [\mathbf{w}_1, \mathbf{w}_2,...,\mathbf{w}_{N_a} ]^\top$ for the graph Laplacian $\mathbf{L}$, we perform change of coordinates $\mathbf{Y} = [\mathbf{y}_1, \mathbf{y}_2,...,\mathbf{y}_{N_\mathrm{a}}]^\top = \mathbf{T}^{-1} \mathbf{X} $ such that
\begin{equation}
    \mathbf{T} \dot{\mathbf{Y}} = -\mathbf{L} \mathbf{T} \mathbf{Y}+ \mathbf{B}.\label{eqn:prop42_1}
\end{equation}
Decompose $\mathbf{L}$ by its right eigenvectors such that $\mathbf{L} = \mathbf{T\Lambda T}^{-1}$, Equation \eqref{eqn:prop42_1} further simplifies as
\begin{align}
    \mathbf{T} \dot{\mathbf{Y}} &= -\mathbf{T\Lambda T}^{-1} \mathbf{T} \mathbf{Y}+ \mathbf{B} = -\mathbf{T\Lambda} \mathbf{Y}+ \mathbf{B}.
\end{align}
Multiplying both sides with $\mathbf{T}^{-1}$ yields
\begin{equation}
    \dot{\mathbf{Y}} = \mathbf{-\Lambda} \mathbf{Y}+ \mathbf{T}^{-1}\mathbf{B}.\label{eqn:prop42_2}
\end{equation}
Since the eigenvalue matrix $\mathbf{\Lambda}$ is diagonal, Equation \eqref{eqn:prop42_2} can be decoupled into
\begin{equation}\label{eq:grand++2_appendix}
    \dot{\mathbf{y}}^\top_i = -\lambda_i \mathbf{y}^\top_i + \mathbf{w}_i^\top \mathbf{B},
\end{equation}
where $\lambda_i$ is the $i$-th eigenvalue of $\mathbf{\Lambda}$. Consider the case of $\lambda_{i}>0$, the solution to the ODE becomes
\begin{equation}
    \mathbf{y}_i^\top (t) = \frac{\mathbf{b}_i}{\lambda_i} + \mathbf{c}^\top_i e^{- \lambda_i t},
\end{equation}
where $\mathbf{b}_i = \mathbf{w}_i^\top \mathbf{B}$ and $\mathbf{c}_i$ are constant vectors.
Consider the case where $\lambda_0 = 0$, the term $ -\lambda_i \mathbf{y}^\top_i$ in Equation \eqref{eq:grand++2_appendix} becomes $\mathbf{0}$, hence the solution to the ODE becomes
\begin{equation}
    \mathbf{y}^\top_0(t) =  \mathbf{b}_0 t + \mathbf{c}^\top_0.
\end{equation}
where $\mathbf{b}_0 = \mathbf{w}_0^\top \mathbf{B} $ and $\mathbf{c}_0$ are constant vectors. 
The solution in the original coordinate frame becomes
\begin{equation}
    \mathbf{X}(t) = \sum_{\lambda_i >0} \mathbf{v}_i \left( \frac{\mathbf{b}_i}{ \lambda_i} + \mathbf{c}^\top_i e^{- \lambda_i t} \right)  + \mathbf{v}_0 (\mathbf{b}_0t + \mathbf{c}^\top_0),\label{eqn:prop41_res}
\end{equation}
where $\lambda_i$ and $\mathbf{v}_i$ denote the positive eigenvalues and corresponding Eigenvectors of $\mathbf{L}$;
$\mathbf{b}_i = \mathbf{w}_i^\top \mathbf{B}$, where $\mathbf{w}_i$ are the left eigenvectors of $\mathbf{L}$. 
Particularly, $\mathbf{v}_0$ is an all-ones vector, i.e., $\mathbf{v}_0 = [1,1,..,1]^\top$, $\mathbf{b}_0 = \mathbf{w}_0^\top \mathbf{B}$, and $\mathbf{c}_0$, $\mathbf{c}_i$ are constant vectors. As $t\rightarrow\infty$, the exponential terms $\mathbf{c}_i^\top e^{-\lambda_i t}$ decays to zero and Equation \eqref{eqn:prop41_res} tends towards a linear system dominated by $\mathbf{v}_0$. Moreover, for large timescales, the difference in node features is relatively small resulting in reduced discriminability (another form of oversmoothing), and poorer network performance.
\end{proof}

\begin{tcolorbox}[blanker, interior engine=spartan, colback=CornflowerBlue!25, boxsep=1mm, borderline west={1.5pt}{0pt}{gray}]
\begin{proposition}[Oversmoothing in GREAD-F and GREAD-FB*]\label{lemma:gread}
    GREAD-F and GREAD-FB${^*}$ exhibits oversmoothing.   
\end{proposition}
\end{tcolorbox}

\begin{proof}
\textbf{GREAD-F.}
In GREAD-F, the layer-wise transformations is of the form,
\begin{equation}\label{eq:gread_f}
    \dot{\mathbf{X}} = -\mathbf{L}\mathbf{X} + \mathbf{X}\odot(1-\mathbf{X}),
\end{equation}
where $\odot$ denotes the Hadamard product. For Laplacian $\mathbf{L}$, there exist a constant $C>0$ such that
\begin{equation}
    |[\mathbf{LX}]_i| \leq C|\mathbf{X}_i|,
\end{equation}
where $C$ depends on the maximum degree and largest edge weights and $\mathbf{X}_i$ denotes the $i$-th row of $\mathbf{X}$. Equation \eqref{eq:gread_f} can therefore be rewritten as
\begin{equation}\label{eq:gread_f_1}
    \dot{\mathbf{X}_i} = - [\mathbf{LX}]_i + \mathbf{X}_i (1-\mathbf{X}_i) \leq C|\mathbf{X}_i|+ \mathbf{X}_i-\mathbf{X}^2_i .
\end{equation}
Notably, when $\mathbf{X}_i<-C$, the RHS of Equation \eqref{eq:gread_f_1} is strictly smaller than zero. Since the dominant term $\mathbf{X}_i^2$ grows quadratically, the solution diverges to infinity
\begin{equation}
    \lim_{t \to \infty} \mathbf{X}_i(t) = -\infty \quad \text{for all } i.
\end{equation}
Therefore, as long as the maximum of $\mathbf{X}$ is smaller than some negative threshold $-C$, the entire system becomes monotonically decreasing, eliminating the possibility of equilibrium or steady-state convergence. While node values may diverge at different rates, the components with the fastest decay rate dominates the overall behavior. The remaining components which decays more slowly becomes negligible in relative magnitude. Therefore, all node features collapse and oversmoothing occurs.

\textbf{GREAD-FB${^*}$.} In GREAD-FB${^*}$, the layer-wise transformations is of the form, 
\begin{equation}
    \dot{\mathbf{X}} = -\alpha\mathbf{L}\mathbf{X} + \beta (\mathbf{LX} +\mathbf{X}),
\end{equation}
where $\alpha,\beta$ are trainable parameters to (de-)emphasize each term. As the external input is a linear transformation, the dynamics can be rewritten as
\begin{equation}
    \dot{\mathbf{X}} = \underbrace{\biggl((\beta -\alpha)\mathbf{L} + \beta \mathbf{I}\biggr)}_{\tilde{\mathbf{L}}}\mathbf{X}.
\end{equation}
Given the property of linear opinion dynamics, when $\alpha > \beta >0$, the eigenvalue of $\tilde{\mathbf{L}}$ exists in range $[2\Delta(\beta-\alpha), \beta]$, where $\Delta$ is the maximum graph degree. In particular, the smallest eigenvalue of $\mathbf{L}$, which is zero, maps to the largest eigenvalue of $\lambda_{\tilde{\mathbf{L}}} = \beta$. This guarantees a global stable equilibrium cannot exist as at least one mode grows exponentially with rate $\beta$. 

Meanwhile, as $t\to\infty$, the remaining components associated with smaller eigenvalues either increases slowly or decays to zero (depends on the sign of the eigenvalue), and thus becoming negligible in relative magnitude. Consequently, all node features are asymptotically dominated by the leading eigenvector, which is $[1,\cdots,1]^\top$. This leads to a collapse in feature diversity and oversmoothing occurs.
\end{proof} 

\subsection{Behavior-Inspired Message Passing Neural Network}
\subsubsection{Lemma \ref{prop:uncorrelated_equation}}
\begin{tcolorbox}[blanker, interior engine=spartan, colback=CornflowerBlue!25, boxsep=1.5mm, borderline west={1.5pt}{0pt}{gray}]
\textit{The general form of nonlinear opinion dynamics (Equation \eqref{eq:BIMP}),
\begin{equation*}
    \dot{\mathbf{X}}(t)
    = -d \mathbf{X}(t)
    + \tanh\left[u \left( \alpha \mathbf{X}(t) + 
    \mathbf{A}^\mathrm{a} \mathbf{X}(t) 
    + \mathbf{X}(t) {\mathbf{A}^\mathrm{o}}^\top +
    \mathbf{A}^\mathrm{a} \mathbf{X}(t) {\mathbf{A}^\mathrm{o}}^\top
    \right) \right] 
    + \mathbf{X}(0),
\end{equation*}
can be written,
\begin{equation}
        \dot{\mathbf{x}} = -d\mathbf{x} + \tanh\left[ u\biggl((\alpha - 1)\mathbf{x} + \tilde{\mathbf{A}} \mathbf{x}\biggr) \right] + \mathbf{b},
    \end{equation}
    where $\tilde{\mathbf{A}}$ is the effective adjacency matrix, $\mathbf{x} = \operatorname{vec}(\mathbf{X})$ and $\mathbf{b} = \operatorname{vec}(\mathbf{B})$.}
\end{tcolorbox}

\begin{proof}
    We first rewrite Equation \eqref{eq:BIMP} as
\begin{equation}
    \dot{\mathbf{X}}
    = -d \mathbf{X} 
    + \tanh\Big[u \Big( (\alpha-1) \mathbf{X} + 
    (\mathbf{A}^\mathrm{a} + \mathbf{I} ) \mathbf{X} ( {\mathbf{A}^\mathrm{o}}^\top + \mathbf{I})
    \Big) \Big] 
    + \mathbf{B}.
\end{equation}
From here, we can write the matrix product \( \mathbf{ABC} \) with \( \mathbf{A} \in \mathbb{R}^{m \times m} \), \(\mathbf{ B} \in \mathbb{R}^{m \times n} \) and \( \mathbf{C} \in \mathbb{R}^{n \times n} \), as \( \mathbf{ABC} = \operatorname{vec}^{-1}\left[\left(\mathbf{C}^\top \otimes \mathbf{A}\right) \operatorname{vec}(\mathbf{B})\right] \), where \( \operatorname{vec} \) denotes the vectorization operator~\cite{magnus2019matrix}. This yields the following form
\begin{equation}
\label{eq:something}
    \dot{\mathbf{X}}
    = -d \mathbf{X} 
    + \tanh\left[u \left( (\alpha-1)\mathbf{X} + \operatorname{vec^{-1}}(
    \tilde{\mathbf{A}} \operatorname{vec}(\mathbf{X})
    \right) \right] 
    + \mathbf{B},
\end{equation}
where $\tilde{\mathbf{A}} = (\mathbf{A}^\mathrm{o}+ \mathbf{I} ) \otimes (\mathbf{A}^\mathrm{a} + \mathbf{I} )$ follows from Definition \ref{def:kroneker}. Vectorizing both sides of yields
\begin{equation}
\label{eq:BINN_vec}
    \dot{\mathbf{x}} = -d \mathbf{x}
    + \tanh\left[u \left( (\alpha-1) \mathbf{x} + 
    \tilde{\mathbf{A}} \mathbf{x}
    \right) \right] 
    + \mathbf{b},
\end{equation}
where $\mathbf{x} = \operatorname{vec}(\mathbf{X})$, and $\mathbf{b} = \operatorname{vec}(\mathbf{B})$. We obtain the general nonlinear opinion dynamics in the form of Equation~\eqref{eq:uncorrelated_options}. By vectorizing $\mathbf{X}$, each agent opinion on an option is treated as an individual agent-like entity, thereby reducing the original dynamics to the form where options are uncorrelated.
\end{proof}

\subsubsection{Lemma \ref{proposition:lambda}}\label{appendix:lambda}
\begin{tcolorbox}[blanker, interior engine=spartan, colback=CornflowerBlue!25, boxsep=1.5mm, borderline west={1.5pt}{0pt}{gray}]
\textit{The leading eigenvalue of the effective adjacency matrix $\tilde{\mathbf{A}}$ is $\lambda^{\tilde{a}}_{\max}=4$.}
\end{tcolorbox}
\begin{proof} 
Since both the communication adjacency matrix $\mathbf{A}^\mathrm{a}$ and the belief adjacency matrix $\mathbf{A}^\mathrm{o}$ are right stochastic, their leading eigenvalues are equal to one (i.e., $\lambda^{\mathrm{a}}_{\max} = \lambda^{\mathrm{o}}_{\max}= 1$). Since $\tilde{\mathbf{A}}=(\mathbf{A}^\mathrm{o} + \mathbf{I})\otimes (\mathbf{A}^\mathrm{a} + \mathbf{I})$, its leading eigenvalue is equal to $\lambda^{\tilde{a}}_{\max} = (\lambda^{\mathrm{a}}_{\max}+1) (\lambda^{\mathrm{o}}_{\max}+1) = 4$.
\end{proof}

\subsubsection{Lemma \ref{proposition:u}: Bifurcation point $u^*$}\label{appendix:u}
\begin{tcolorbox}[blanker, interior engine=spartan, colback=CornflowerBlue!25, boxsep=1.5mm, borderline west={1.5pt}{0pt}{gray}]
\textit{The bifurcation point $u^*$ of the attention-opinion bifurcation diagram is equal to $\nicefrac{d}{\left(\alpha-1 + \lambda^{\tilde{a}}_{\max}\right)}$.}
\end{tcolorbox}

\begin{proof} 
The bifurcation point of the attention-opinion bifurcation diagram, i.e., the point where the number and/or stability of the solutions change,  occurs when the input parameter $\mathbf{B}$ is equal to zero.

Following \cite{leonard2024fast}, we use a linear analysis to determine the bifurcation point of the attention-opinion bifurcation diagram. The linearization of the nonlinear opinion dynamics model (Equation \eqref{eq:BINN_vec}) is given by $\dot{\boldsymbol{\omega}} = J(\mathbf{x}_e)\,\boldsymbol{\omega}$, where $J(\mathbf{x}_e)$ is the Jacobian evaluated at the equilibrium $\mathbf{x}_e$, and $\boldsymbol{\omega} = \mathbf{x}-\mathbf{x}_e$. At the equilibrium $\mathbf{x}_e=\mathbf{0}$, the Jacobian is given by 
\begin{equation}
\label{eq:jacobian}
    \mathbf{J} = ( -d+u(\alpha-1))\mathbf{I} + u\tilde{\mathbf{A}}.
\end{equation}
The eigenvalue of the Jacobian determines the equilibrium  stability. Denoting the eigenvalue of $\tilde{\mathbf{A}}$, $\lambda^{\tilde{a}}$, the eigenvalue of the Jacobian can be expressed,
\begin{equation}
    \lambda'=-d+u(\alpha-1)+u\lambda^{\tilde{a}}.
\end{equation}

The bifurcation point of the attention-opinion bifurcation diagram occurs when the dominant eigenvalue of the Jacobian is zero, reaching the upper bound for stability of the equilibrium $\mathbf{x}_e$. As $u$ continues to increases and the dominant eigenvalue becomes positive, the equilibrium $\mathbf{x}_e$ become unstable and a bifurcation emerges. Solving for the critical value of the attention yields $u^* = \nicefrac{d}{\left(\alpha-1 + \lambda^{\tilde{a}}_{\max}\right)}$.
\end{proof}

\subsubsection{Lemma \ref{proposition:stability}: BIMP converges to equilibrium}\label{appendix:stability}
\begin{tcolorbox}[blanker, interior engine=spartan, colback=CornflowerBlue!25, boxsep=1.5mm, borderline west={1.5pt}{0pt}{gray}]
\textit{A BIMP model with time-independent input parameter $\mathbf{B}$, converges to an equilibrium.}
\end{tcolorbox}
\begin{proof}
Due to the monotonicity of our BIMP model, the opinions are guaranteed to converge to an equilibrium. 
Without loss of generality, consider the case where the graph is undirected and the system has only one option (i.e., $\mathbf{A}^{\mathrm{o}}=0$ with no interrelationship between options),
\begin{equation}\label{eq:binn_stability}
    \dot{\mathbf{X}} = -d\mathbf{X} + \operatorname{tanh}\biggl[ u (\mathbf{A^\mathrm{a}}\mathbf{X}  + \alpha \mathbf{X}) \biggr] + \mathbf{B}.
\end{equation}
Let $\mathbf{p}$ be an permutation matrix that re-index our system into block diagonal form
\begin{align}
    \hat{\mathbf{A}}^\mathrm{a} &= \mathbf{P}\mathbf{A^\mathrm{a}}\mathbf{P}^\top = 
    \begin{bmatrix}
    \mathbf{A}_{11} & 0 & 0 & 0 \\
    0 & \mathbf{A}_{22} & 0 & 0 \\
    \vdots & \vdots & \ddots & 0 \\
    0 & 0 & \dots & \mathbf{A}_{nn} 
    \end{bmatrix},\label{eq:block_diagonal}\;\;\;
    \hat{\mathbf{X}} = \mathbf{P}\mathbf{X} = 
    \begin{bmatrix}
    \mathbf{X}_{1}\\
    \mathbf{X}_{2}\\
    \vdots\\
    \mathbf{X}_{n}
    \end{bmatrix},\;\;\;
    \hat{\mathbf{B}} = \mathbf{P}\mathbf{B} = 
    \begin{bmatrix}
    \mathbf{B}_{1}\\
    \mathbf{B}_{2}\\
    \vdots\\
    \mathbf{B}_{n}
    \end{bmatrix},
\end{align}
where $\mathbf{A}_{nn}$ are irreducible blocks or zero matrices. Considering $\mathbf{p}$ is an orthonormal vector, $\mathbf{A^\mathrm{a}}$ and $\mathbf{X}$ can be expressed as
\begin{equation}
    \mathbf{A^\mathrm{a}} = \mathbf{P}^\top \hat{\mathbf{A}}^\mathrm{a} \mathbf{P}, \quad \mathbf{X} = \mathbf{P}^\top \hat{\mathbf{X}}, \quad \mathbf{B} = \mathbf{P}^\top \hat{\mathbf{B}}.
\end{equation}
Substituting $\mathbf{A^\mathrm{a}}$ and $\mathbf{X}$ with $\hat{\mathbf{A}}^\mathrm{a}$ and $\hat{\mathbf{X}}$ respectively in Equation \eqref{eq:binn_stability} yields
\begin{align}
    \mathbf{P}^\top \dot{\hat{\mathbf{X}}} &= -d\mathbf{P}^\top \hat{\mathbf{X}} + \operatorname{tanh}\biggl[u( \mathbf{P}^\top \hat{\mathbf{A}}^\mathrm{a} \mathbf{PP^\top} \hat{\mathbf{X}}  + \alpha \mathbf{P}^\top \hat{\mathbf{X}}) \biggr] +  \mathbf{P}^\top \hat{\mathbf{B}},\\
     &= \mathbf{P}^\top (-d\hat{\mathbf{X}}) + \mathbf{P}^\top \operatorname{tanh}\biggl[u( \hat{\mathbf{A}}^\mathrm{a} \hat{\mathbf{X}}  + \alpha \hat{\mathbf{X}} ) \biggr]  +  \mathbf{P}^\top \hat{\mathbf{B}},
\end{align}
and multiplying $\mathbf{P}$ on both sides
\begin{equation}\label{eq:binn_permuted}
    \dot{\hat{\mathbf{X}}} =  -d\hat{\mathbf{X}} + \operatorname{tanh}\biggl[ u( \hat{\mathbf{A}}^\mathrm{a} \hat{\mathbf{X}}  + \alpha \hat{\mathbf{X}} ) \biggr]  + \hat{\mathbf{B}}.
\end{equation}
Leveraging the block diagonal form, Equation \eqref{eq:binn_permuted} can be decoupled into 
\begin{equation}\label{eq:binn_components}
    \dot{\mathbf{X}}_m = -d\mathbf{X}_m + \operatorname{tanh}\biggl[ u ( \mathbf{A}_{mm}\mathbf{X}_m + \alpha \mathbf{X}_m )\biggr] + \mathbf{B}_m.
\end{equation}
where the convergence of each subsystem $\dot{\mathbf{X}}_m = f_m(\mathbf{X}_m), 1 \leq m\leq n$ can be examined individually.
The Jacobian of subsystem $\dot{\mathbf{X}}_m = f_m(\mathbf{X}_m)$ is defined as
\begin{equation}
    \mathbf{J}_m(\mathbf{X}_m) = \frac{\partial f_m(\mathbf{X}_m) }{\partial \mathbf{X}_m} =
    -d\mathbf{I} + \mathbf{1}\text{vec}\biggl(\operatorname{sech}^2\left( u( \mathbf{A}_{mm}\mathbf{X}_m + \alpha \mathbf{X}_m ) \right)\biggl)^{\top} \circ \left[ u ( \left(\mathbf{A}_{mm} \otimes \mathbf{I} \right) + \alpha \mathbf{I}) \right],
\end{equation}
where $\mathbf{I}$ is the identity matrix, $\mathbf{1}$ the all-ones vector and $\text{vec}$ is the vectorization. $\circ$ is Hadamard product. Each subsystem and their associated Jacobian satisfies

\begin{itemize}
    \itemsep0em 
    \item \textbf{Cooperative.} Since \(\operatorname{sech} \in (0,1]\), \(\mathbf{A}_{mm}\) are positive matrices (as it is the output of a softmax function) and \(\alpha\geq 0\), \(\mathbf{J}_m(\mathbf{X}_m)\) is a Metzler matrix in which all the off-diagonal elements are non-negative. This implies that \(\mathbf{J}_m(\mathbf{X}_m)\) is \textit{cooperative} and the subsystem \(\dot{\mathbf{X}}_m =f_m(\mathbf{X}_m)\) is monotone~\cite{hirsch1985systems}.
    \item \textbf{Irreducible.} As \( \mathbf{A}_{mm} \) is constructed to be irreducible, \(((\mathbf{A}_{mm} \otimes \mathbf{I}) +  \alpha \mathbf{I})\) remains irreducible and hence the Jacobian \(\mathbf{J}_m(\mathbf{X}_m)\) is \textit{irreducible}.
    \item \textbf{Compact closure.} The existence of the damping term $d$ in subsystem $\dot{\mathbf{X}}_m = f_m(\mathbf{X}_m)$ ensure the forward orbit has compact closure (i.e., bounded).
\end{itemize}
If the Jacobian for a continuous dynamical system $\dot{x}=f(x,t)$ is cooperative and irreducible, then it approaches the equilibrium for almost every point $x$ whose forward orbit has compact closure \cite{hirsch1985systems}. 
Since the Jacobian $\mathbf{J}_{m}(\mathbf{X}_{m})$ for each subsystem $\dot{\mathbf{X}}_m = f_m(\mathbf{X}_m)$ satisfies this condition, almost every state $\mathbf{X}_{m}$ approaches the equilibrium set. 
Therefore the dynamical system defined in Equation \eqref{eq:binn_stability} converges to an equilibrium set. 
As all trajectories tend towards the equilibrium solution, analyzing the equilibrium behavior is sufficient to understand the dynamics of BIMP.

If there are more than one option in the system (i.e, $\mathbf{A}^\mathrm{o} \neq [0]$), the vectorized system defined in Equation \eqref{eq:BINN_vec} can be shown analogously to converge to its equilibrium set.
\end{proof}

\subsubsection{Theorem \ref{theorem:b!=0}: Dissensus in BIMP}
\begin{tcolorbox}[blanker, interior engine=spartan, colback=CornflowerBlue!25, boxsep=1.5mm, borderline west={1.5pt}{0pt}{gray}]
\textit{BIMP will not exhibit oversmoothing when the input parameter $\mathbf{B}$ is time-independent with unique entries.}
\end{tcolorbox}

\begin{proof}
Without loss of generality, consider the case where the graph is undirected and the system has only one option (i.e., $\mathbf{A}^{\mathrm{o}}=\mathbf{0}$). By Definition \ref{def:kroneker}, the effective adjacency matrix becomes
\begin{equation}
    \tilde{\mathbf{A}} = 1\otimes (\mathbf{A}^\mathrm{a}+\mathbf{I}) = \mathbf{A}^\mathrm{a}+\mathbf{I}.
\end{equation}
Consider that $\mathbf{x}= [x_1, x_2, ..., x_{N_a}]^\top $, we can decouple the dynamical equation of $x_i$ from Equation \eqref{eq:BINN_vec} such that
\begin{equation}
    \dot{x}_i = -dx_i + \operatorname{tanh} (u(\tilde{\alpha} x_i + \tilde{\mathbf{a}}_i \mathbf{x})) + b_i,
\end{equation}
where $\tilde{\mathbf{a}}_i$ is the $i$-th row of $\tilde{\mathbf{A}}$ and $b_i$ is the $i$-th element of $\mathbf{b}$. Assume $\mathbf{x}$ converges to consensus such that $x_1 = x_2=\ldots=x_{N_a}=\bar{x}$.
For any pair $x_m$ and $x_n, m\neq n$ with corresponding input $b_m \neq b_n$, their dynamical equations are
\begin{gather}
    \dot{x}_m = -dx_m + \operatorname{tanh} (u(\tilde{\alpha} x_m + \tilde{\mathbf{a}}_m \mathbf{x})) + b_m,\label{eq:x_m}
    \\
    \dot{x}_n = -dx_n + \operatorname{tanh} (u(\tilde{\alpha} x_n + \tilde{\mathbf{a}}_n \mathbf{x})) + b_n.\label{eq:x_n}
\end{gather}
We observe that
\begin{gather}
    -dx_{m}=-dx_{n}=-d\bar{x},\\
    -\tilde{\alpha}x_{m}=-\tilde{\alpha}x_{n}=-\tilde{\alpha}\bar{x},
\end{gather}
and
\begin{equation}
    \tilde{\mathbf{a}}_m \mathbf{x} =  \tilde{\mathbf{a}}_n \mathbf{x} = 2\bar{x},
\end{equation}
due to $\tilde{\mathbf{a}}_m$ and $\tilde{\mathbf{a}}_n$ being right stochastic. However, since $b_m \neq b_n$, the right hand side of Equation \eqref{eq:x_m} and Equation \eqref{eq:x_n} cannot be $0$ at the same time. Therefore, by contradiction, consensus cannot be the equilibrium for BIMP if $b_m \neq b_n$. If $\mathbf{b}$ has unique elements, the equilibrium of the system forms dissensus and avoids oversmoothing.

If there are more than one option in the system (i.e, $\mathbf{A}^o \neq \mathbf{0}$), formation of dissensus is still possible since $\mathbf{A}^{\mathrm{o}}$ is also right stochastic.
\end{proof}

\subsubsection{Theorem \ref{theorem:gradient}: BIMP has well behaved gradients}\label{appendix:gradient}
\begin{tcolorbox}[blanker, interior engine=spartan, colback=CornflowerBlue!25, boxsep=1.5mm, borderline west={1.5pt}{0pt}{gray}]
\textit{BIMP gradients are upper bounded and do not vanish exponentially.}
\end{tcolorbox}

\begin{proof}
For simplicity, consider the forward Euler method for integration of the dynamics defined by Equation \eqref{eq:BIMP} such that
\begin{gather}
    \mathbf{X}^t = \mathbf{X}^{t-1} + \Delta t \dot{\mathbf{X}}^{t-1},\label{eqn:euler1}\\
    \mathbf{X}^0 = \phi(\mathbf{X_{\operatorname{in}}}) = \mathbf{W}\mathbf{X}_{\operatorname{in}},\label{eqn:euler2}
\end{gather}
where $\Delta t$ is the numerical integration step size, $\mathbf{X}^{t}\in \mathbb{R}^{N_a\times N_o}$ are the features at time $t\in\left[\Delta t, 2\Delta t ,\ldots, M\Delta t\right]$, and $\mathbf{X}_{\operatorname{in}}\in \mathbb{R}^{N_a\times f}$ are the input features. For the simplicity, we assume a linear encoder $\phi$ parameterized by learnable weights $\mathbf{W}$. Similar to existing continuous depth GNNs, the total steps of ODE integrations $M$ is interpreted as the number of layers of a model. Consider a node classification task using BIMP subject to mean squared error loss
\begin{equation}\label{eq:loss_function}
    \mathcal{L}(\mathbf{W}) = \frac{1}{2N_\mathrm{a}N_\mathrm{o}}\sum_{i=1}^{N_\mathrm{a}}\sum_{j=1}^{N_\mathrm{o}}(x_{ij}^M - \hat{x}_{ij})^2,
\end{equation}
where $x^{M}_{ij}$ is an element of the learned features $\mathbf{X}^{M}$ at layer $M$ and $\hat{x}_{ij}$ is an element of the ground truth $\hat{\mathbf{X}}$. Consider all intermediate layers where $t \in [\Delta t, 2\Delta t, \ldots , M\Delta t]$, the gradient descent equation can be constructed as
\begin{equation}\label{eq:chain_rule}
    \frac{\partial \mathcal{L}}{\partial \mathbf{W}} = \frac{\partial \mathcal{L}}{\partial \mathbf{X}^M} \frac{\partial \mathbf{X}^M}{\partial \mathbf{X}^1} \frac{\partial \mathbf{X}^1}{\partial \mathbf{X}^0} \frac{\partial \mathbf{X}^0 }{\partial \mathbf{W}},
\end{equation}
where 
\begin{equation}
    \frac{\partial \mathbf{X}^M}{\partial \mathbf{X}^1} = \prod_{t=2}^M \frac{\partial \mathbf{X}^t}{\partial \mathbf{X}^{t-1}}.
\end{equation}
With increasing depth (i.e, $M \rightarrow \infty$), this repeated multiplication leads to gradient exploding (vanishing) when the components $\nicefrac{\partial \mathbf{X}^t}{\partial \mathbf{X}^{t-1}} > \mathbf{I}$ ($\nicefrac{\partial \mathbf{X}^t}{\partial \mathbf{X}^{t-1}} <\mathbf{I}$). The BIMP model provides an upper and lower bound on gradients in Lemma \ref{proposition:gradient_explode} and \ref{proposition:gradient_vanish} to guarantee exploding or vanish gradients cannot occur. 
\end{proof}

\begin{tcolorbox}[blanker, interior engine=spartan, colback=CornflowerBlue!25, boxsep=1mm, borderline west={1.5pt}{0pt}{gray}]
\begin{proposition}
\label{proposition:gradient_explode}
    BIMP gradients are upper bounded when the step-size $\Delta t \ll 1$ and damping term $d<\nicefrac{1}{\Delta t}$.
\end{proposition}
\end{tcolorbox}

\begin{proof}
Consider integrating BIMP with the forward Euler scheme defined in Equation \eqref{eqn:euler1} and \eqref{eqn:euler2} with fixed hyper parameters $\tilde{\alpha} =\alpha-1$ and $u = \nicefrac{d}{\tilde{\alpha}+4}$,
\begin{align}
    \mathbf{X}^t &= \mathbf{X}^{t-1} + \Delta t \biggl( -d\mathbf{X}^{t-1} 
    + \tanh\biggl[ u \biggl( \tilde{\alpha} \mathbf{X}^{t-1} + 
    \bigl(\mathbf{A}^\mathrm{a} + \mathbf{I}\bigr) \mathbf{X}^{t-1} \bigl(\mathbf{A}^\mathrm{o\top} + \mathbf{I}\bigr)
    \biggr)
    \biggr] 
    + \mathbf{X}^0
    \biggr) \\
    &= (1-d\Delta t) \mathbf{X}^{t-1} 
    + \Delta t \tanh \biggl[ u  \biggl( \tilde{\alpha} \mathbf{X}^{t-1} +
    \bigl(\mathbf{A}^\mathrm{a} + \mathbf{I}\bigr) \mathbf{X}^{t-1} \bigl(\mathbf{A}^\mathrm{o\top} + \mathbf{I}\bigr)
    \biggr)
    \biggr]
    + \Delta t \mathbf{X}^0, \label{eq:Euler_update}
\end{align}
with initial feature embedding
\begin{equation}
    \mathbf{X}^0 = \phi\bigl(\mathbf{X}_{\operatorname{in}}\bigr) = \mathbf{X}_{\operatorname{in}} \mathbf{W}.\label{eq:Euler_update_2}
\end{equation}
Vectorizing Equation \eqref{eq:Euler_update} and \eqref{eq:Euler_update_2} yields
\begin{gather}
    \mathbf{x}^t = (1-d\Delta t) \mathbf{x}^{t-1} + \Delta t \tanh\left[u \biggl( \tilde{\alpha}\mathbf{I} + \tilde{\mathbf{A}}\biggr) \mathbf{x}^{t-1}\right] + \Delta t \mathbf{x}^0,\label{eq:Euler_update_vectorized}\\
    \mathbf{x}^0 = \tilde{\mathbf{W}} \mathbf{x}_{\operatorname{in}},
\end{gather}
where $\tilde{\mathbf{A}} = (\mathbf{A}^\mathrm{o}+ \mathbf{I} ) \otimes (\mathbf{A}^\mathrm{a} + \mathbf{I} )$, $\tilde{\mathbf{W}}=\mathbf{W}^{\top}\otimes \mathbf{I}_{N_a}$, and $\mathbf{x}^t = [x^t_1, x^t_2,\ldots, x^t_{N_\mathrm{a}\times N_\mathrm{o}}]^{\top}$. Reformulating gradient calculation in Equation \eqref{eq:chain_rule} subject to loss function defined in Equation \eqref{eq:loss_function} with respect to the vectorized variables gives
\begin{gather}
    \frac{\partial \mathcal{L}}{\partial \tilde{\mathbf{W}}} =\frac{\partial \mathcal{L}}{\partial \mathbf{x}^M} \frac{\partial \mathbf{x}^M}{\partial \mathbf{x}^1} \frac{\partial \mathbf{x}^1}{\partial \mathbf{x}^0} \frac{\partial \mathbf{x}^0 }{\partial \tilde{\mathbf{W}}},\\
    \frac{\partial \mathbf{x}^M}{\partial \mathbf{x}^1} = \prod_{t=2}^M \frac{\partial \mathbf{x}^t}{\partial \mathbf{x}^{t-1}},
\end{gather}
where the upper bound for $\left\|\frac{\partial \mathbf{x}^M}{\partial \mathbf{x}^1}\right\|_\infty $, $\left\|\frac{\partial \mathcal{L}}{\partial \mathbf{x}^M} \right\|_\infty$, $\left\|\frac{\partial \mathbf{x}^1}{\partial \mathbf{x}^0}\right\|_\infty $ and $\left\|\frac{\partial \mathbf{x}^0}{\partial \tilde{\mathbf{W}}}\right\|_\infty $ can be found individually and are summarized in Equation \eqref{eqn:bound_1_final}, \eqref{eqn:bound_2_final}, \eqref{eq:z1/z0}, and \eqref{eq:z0/w} respectively.

Consider the first term $\left\|\frac{\partial \mathbf{x}^M}{\partial \mathbf{x}^1}\right\|_\infty $ and recalling that 
\begin{equation}
    \frac{\partial \mathbf{x}^M}{\partial \mathbf{x}^1} = \prod_{t=2}^M \frac{\partial \mathbf{x}^t}{\partial \mathbf{x}^{t-1}}.
\end{equation}
By inspecting each term $\frac{\partial \mathbf{x}^t}{\partial \mathbf{x}^{t-1}}$, it follows that
\begin{equation}
    \frac{\partial \mathbf{x}^t}{\partial \mathbf{x}^{t-1}} = (1-d\Delta t)\mathbf{I}
    + \Delta t \mathbf{1} \left[\operatorname{sech}^2 \left( u ( \tilde{\alpha}\mathbf{I} + \tilde{\mathbf{A}} ) \mathbf{x}^{t-1} \right) \right]^{\top} \circ \biggl( u (\tilde{\alpha}\mathbf{I} + \tilde{\mathbf{A}} ) \biggr),
\end{equation}
where $\mathbf{1} \left[\operatorname{sech}^2 \left( u ( \tilde{\alpha}\mathbf{I} + \tilde{\mathbf{A}} ) \mathbf{x}^{t-1} \right) \right]^{\top}$ represents a matrix repeating the vector $\operatorname{sech}^2 \left( u ( \tilde{\alpha}\mathbf{I} + \tilde{\mathbf{A}} ) \mathbf{x}^{t-1} \right)$ 
 along the row dimension. $\circ$ is the Hadamard product. As $\operatorname{sech}(\cdot) \in (0,1]$, we can leverage the triangle identity to obtain an upper bound 
\begin{align}
    \left\| \frac{\partial \mathbf{x}^t}{\partial \mathbf{x}^{t-1}}\right\|_\infty 
    &\leq \left\|(1-d\Delta t) \mathbf{I} + u \Delta t (\tilde{\alpha}\mathbf{I} +\tilde{\mathbf{A}}) \right\|_\infty\\
    &\leq \left\|(1- d\Delta t) \mathbf{I} \right\|_\infty + u \Delta t \left\| \tilde{\alpha}\right\|_\infty + u \Delta t \left\|  \tilde{\mathbf{A}} \right\|_\infty.\label{eq:upper_bound_a}
\end{align}
Since $u=\nicefrac{d}{\alpha+3}, \tilde{\alpha} = \alpha-1, d>0, \alpha\geq0$, it follows that
\begin{equation}
    u\Delta t \left\| \tilde{\alpha}\right\|_\infty = \frac{d}{\alpha+3}\Delta t \left\| \alpha - 1 \right\| < d\Delta t.\label{eqn:upper_bound_a_1}
\end{equation}
Since $\tilde{\mathbf{A}} = (\mathbf{A}^\mathrm{o}+ \mathbf{I} ) \otimes (\mathbf{A}^\mathrm{a}  + \mathbf{I} )$ from Definition \ref{def:kroneker} and given that $\mathbf{A}^{a}$ and $\mathbf{A}^{o}$ are right stochastic, it follows that
\begin{equation}
    \left\|  \tilde{\mathbf{A}} \right\|_\infty < 4.\label{eqn:upper_bound_a_2}
\end{equation}
Therefore, Equation \eqref{eq:upper_bound_a} can be further bounded by Equation \eqref{eqn:upper_bound_a_1} and \eqref{eqn:upper_bound_a_2} as
\begin{align}
    \left\| \frac{\partial \mathbf{x}^t}{\partial \mathbf{x}^{t-1}}\right\|_\infty 
    &\leq \left\|(1- d\Delta t) \mathbf{I} \right\|_\infty + d\Delta t +4u\Delta t,\\
    & <  (1-d\Delta t) + d\Delta t +4u\Delta t,\\
    & <  1 + 4u\Delta t.
\end{align}
Since we assume $\Delta t\ll 1$, it follows that
\begin{align}
    \left( 1+ 4u\Delta t \right)^{M-1} &= 1 + 4(M-1)u\Delta t + \mathcal{O}(\Delta t^2),\\
    &\approx 1 + 4(M-1)u\Delta t,\\
    &< 1 + 4Mu\Delta t.
\end{align}
Finally, the term is upper bounded as
\begin{equation}
    \left\| \frac{\partial \mathbf{x}^M}{\partial \mathbf{x}^{1}}\right\|_\infty 
    \leq 1 + 4Mu\Delta t.\label{eqn:bound_1_final}
\end{equation}

Consider the second term $\left\|\frac{\partial \mathcal{L}}{\partial \mathbf{x}^M} \right\|_\infty$
\begin{equation}
    \frac{\partial \mathcal{L}}{\partial \mathbf{x}^M} = \frac{1}{N_a N_o} \operatorname{diag}(\mathbf{x}^M - \hat{\mathbf{x}}), \label{eq:j/zm}\\
\end{equation}
where $\operatorname{diag}(\mathbf{x}^M - \hat{\mathbf{x}})$ is a diagonal matrix with vector entry $\mathbf{x}^M - \hat{\mathbf{x}}$ on the diagonal and $\hat{\mathbf{x}}=\operatorname{vec}(\hat{\mathbf{X}})$. Taking the absolute value of Equation \eqref{eq:Euler_update_vectorized} and recalling $\tanh(\cdot) \in (-1,1)$, it follows that
\begin{equation}
    |x_{i}^M| \leq (1-d\Delta t)|x_{i}^{M-1}| + (1 + |x_{i}^0|) \Delta t.
\end{equation}
Therefore, recursively it can be shown that
\begin{align}
         |x_{i}^M| 
        &\leq (1-d\Delta t)^M |x_{i}^0| + \left(\sum_{p=0}^{M-1} (1-d\Delta t)^p \right) (1 + |x_{i}^0|) \Delta t,\\
        &\leq |x_{i}^0| + M (1 + |x_{i}^0|) \Delta t.\label{eq:|zm|}
\end{align}
Taking $\infty$-norm of Equation \eqref{eq:j/zm} yields
\begin{equation}\label{eq:|j/zm|}
    \left\|\frac{\partial \mathcal{L}}{\partial \mathbf{x}^M} \right\|_\infty 
    \leq 
    \frac{1}{N_a N_o}(\|\mathbf{x}^M \|_\infty+ \|\hat{\mathbf{x}} \|_\infty).
\end{equation}
Substituting Equation \eqref{eq:|zm|} into Equation \eqref{eq:|j/zm|} result in the upper bound
\begin{equation}
    \left\|\frac{\partial \mathcal{L}}{\partial \mathbf{x}^M} \right\|_\infty 
    \leq 
    \frac{1}{N_a N_o}\left( M\Delta t +  (1+M\Delta t ) \|\mathbf{x}^0 \|_\infty + \|\hat{\mathbf{x}} \|_\infty \right).\label{eqn:bound_2_final}
\end{equation}

Consider the third term $\left\|\frac{\partial \mathbf{x}^1}{\partial \mathbf{x}^0}\right\|_\infty$. Since the input term $\mathbf{x}^{0}$ contributes to the differential defined in Equation \eqref{eq:Euler_update}, it follows that the upper bound can be derived as
\begin{gather}
    \frac{\partial \mathbf{x}^1}{\partial \mathbf{x}^0}  = \biggl(1-(d-1)\Delta t \biggr)\mathbf{I} + \Delta t \mathbf{1} \left[\operatorname{sech}^2\left( u ( \tilde{\alpha}\mathbf{I} + \tilde{\mathbf{A}}) \mathbf{x}^{0} 
    \right) \right]^{\top} \circ
    \left( u ( \tilde{\alpha}\mathbf{I} + \tilde{\mathbf{A}}) \right),\\
    \left\|\frac{\partial \mathbf{x}^1}{\partial \mathbf{x}^0} \right\|_\infty < 1 + (4u + 1)\Delta t. \label{eq:z1/z0}
\end{gather}

Consider the fourth term $\left\|\frac{\partial \mathbf{x}^0}{\partial \tilde{\mathbf{W}}}\right\|_\infty$, the upper bound can be defined as
\begin{equation}\label{eq:z0/w}
    \left\| \frac{\partial \mathbf{x}^0}{\partial \tilde{\mathbf{W}}} \right\|_\infty = \left\| \mathbf{x}_{\operatorname{in}} \right\|_\infty.
\end{equation}

Combining Equation \eqref{eqn:bound_1_final}, \eqref{eqn:bound_2_final}, \eqref{eq:z1/z0}, and \eqref{eq:z0/w}, it follows that the upper bound for gradient calculations of BIMP is
\begin{equation}\label{eqn:gradient_upper_bound_overall}
    \left\|\frac{\partial \mathcal{L}}{\partial \tilde{\mathbf{W}}}\right\|_{\infty} 
    < 
    \frac{1}{N_a N_o}\biggl( M\Delta t +  (1+M\Delta t ) \|\mathbf{x}^0 \|_\infty + \|\hat{\mathbf{x}} \|_\infty \biggr)
    \biggl( 1 + 4Mu \Delta t \biggr)
    \biggl(1+(4u+1)\Delta t \biggr) 
    \left\| \mathbf{x}_{\operatorname{in}} \right\|_\infty.
\end{equation}
By designing hyperparameters
\begin{gather}
    \beta = M\Delta t, \\
    \gamma = \left( 1 + 4Mu \Delta t \right)
    \left(1+(4u+1)\Delta t \right),
\end{gather}
the upper bound defined in Equation \eqref{eqn:gradient_upper_bound_overall} can be simplified as
\begin{equation}
    \left\|\frac{\partial \mathcal{L}}{\partial \tilde{\mathbf{W}}}\right\|_{\infty} 
    < 
    \frac{1}{N_a N_o}\left( \beta +  (1+\beta ) \|\mathbf{x}^0 \|_\infty + \|\hat{\mathbf{x}} \|_\infty \right)
    \gamma
    \left\| \mathbf{x}_{\operatorname{in}} \right\|_\infty.
\end{equation}
Consider that $\mathbf{W}$ and $\tilde{\mathbf{W}}$ have the same elements, $\| \frac{\partial \mathcal{L}}{\partial \mathbf{W}} \|_\infty = \| \frac{\partial \mathcal{L}}{\partial \tilde{\mathbf{W}}} \|_\infty$ and therefore
\begin{equation}
    \left\|\frac{\partial \mathcal{L}}{\partial \mathbf{W}}\right\|_{\infty} 
    < 
    \frac{1}{N_a N_o}\left( \beta +  (1+\beta ) \|\mathbf{x}^0 \|_\infty + \|\hat{\mathbf{x}} \|_\infty \right)
    \gamma
    \left\| \mathbf{x}_{\operatorname{in}} \right\|_\infty.
\end{equation}
which indicates that the gradients are upper bounded and avoids exploding gradients.
\end{proof}

\begin{tcolorbox}[blanker, interior engine=spartan, colback=CornflowerBlue!25, boxsep=1mm, borderline west={1.5pt}{0pt}{gray}]
\begin{proposition}
\label{proposition:gradient_vanish}
     BIMP gradients will not vanish exponentially when the step-size $\Delta t \ll 1$ and the damping term $d<\nicefrac{1}{\Delta t}$.
\end{proposition}
\end{tcolorbox}

\begin{proof}
From Equation \eqref{eq:chain_rule}, the terms $\frac{\partial \mathbf{X}^M}{\partial \mathbf{X}^1}$, $\frac{\partial \mathbf{X}^1}{\partial \mathbf{X}^0}$, $\frac{\partial \mathcal{L}}{\partial \mathbf{X}^M}$,  and $\frac{\partial \mathbf{X}^0}{\partial \mathbf{W}}$ can be individually reformulated as a recursive summation operation and are summarized in Equation \eqref{eqn:vanish_1}, \eqref{eqn:vanish_2}, \eqref{eq:j/m}, and \eqref{eq:x0/w} respectively.

Consider the term $\frac{\partial \mathbf{X}^M}{\partial \mathbf{X}^1}$, which can be expressed as
\begin{equation}
    \frac{\partial \mathbf{X}^t}{\partial \mathbf{X}^{t-1}} = \mathbf{I} 
    + \Delta t \mathbf{E}_{t-1},\label{eqn:vanish_1}
\end{equation}
where
\begin{multline}
    \mathbf{E}_{t-1} = -d\mathbf{I} 
    + \mathbf{1} \left[\operatorname{sech}^2\left(u(\tilde{\alpha} \mathbf{X}^{t-1} + (\mathbf{A}^a+\mathbf{I})\mathbf{X}^{t-1} ({\mathbf{A}^o}^{\top}+\mathbf{I})
    \right) \right]^{\top} \\
    \circ  u\biggl[\tilde{\alpha} \mathbf{I} + \biggl(\mathbf{I}\otimes(\mathbf{A}^o+\mathbf{I})\biggr) \biggl( (\mathbf{A}^a+\mathbf{I})\otimes \mathbf{I} \biggr) \biggr].
\end{multline}

Consider the term $\frac{\partial \mathbf{X}^1}{\partial \mathbf{X}^{0}}$, which can be reformulated as
\begin{equation}
    \frac{\partial \mathbf{X}^1}{\partial \mathbf{X}^{0}} = \mathbf{I} 
    + \Delta t \mathbf{E}_{0},\label{eqn:vanish_2}
\end{equation}
where
\begin{multline}
    \mathbf{E}_{0} =
     (1-d)\mathbf{I} 
    + \mathbf{1} \text{vec}\biggl(\operatorname{sech}^2\left(u(\tilde{\alpha} \mathbf{X}^{t-1} + (\mathbf{A}^a+\mathbf{I})\mathbf{X}^{t-1} ({\mathbf{A}^o}^{\top}+\mathbf{I})
    \right) \biggl)^{\top} \\
    \circ  u\biggl[\tilde{\alpha} \mathbf{I} + \biggl(\mathbf{I}\otimes(\mathbf{A}^o+\mathbf{I})\biggr) \biggl( (\mathbf{A}^a+\mathbf{I})\otimes \mathbf{I} \biggr) \biggr].
\end{multline}
Combining the previous two terms, it follows that
\begin{align}
    \frac{\partial \mathbf{X}^M}{\partial \mathbf{X}^{0}} 
    &= \biggl( \mathbf{I} + \Delta t \mathbf{E}_{M-1} \biggr)  \biggl( \mathbf{I} + \Delta t \mathbf{E}_{M-2} \biggr) ... \biggl( \mathbf{I} + \Delta t \mathbf{E}_{0} \biggr),\\
    &= \mathbf{I}_{n} 
    + \Delta t \left(\mathbf{E}_{0} + \sum_{t=1}^{M-1}\mathbf{E}_{t} \right) + \mathcal{O}(\Delta t^2).\label{eq:lower_bound}
\end{align}
Consider the term $\frac{\partial \mathcal{L}}{\partial \mathbf{X}^M}$
\begin{equation}\label{eq:j/m}
    \frac{\partial \mathcal{L}}{\partial \mathbf{X}^M} = \frac{1}{N_a N_o} \operatorname{diag}(\mathbf{X}^M - \hat{\mathbf{X}}),
\end{equation}
where $\operatorname{diag}(\mathbf{X}^M - \hat{\mathbf{X}})$ is a diagonal matrix with vector entry $\operatorname{vec}\left(\mathbf{X}^M - \hat{\mathbf{X}}\right)$ on the diagonal.

Consider the term $\frac{\partial \mathbf{X}^0}{\partial \mathbf{W}}$
\begin{equation}\label{eq:x0/w}
    \frac{\partial \mathbf{X}^0}{\partial \mathbf{W}} = \mathbf{X}_{\operatorname{in}} \otimes \mathbf{I}.
\end{equation}
Therefore, combining Equation \eqref{eq:lower_bound}, \eqref{eq:j/m} and \eqref{eq:x0/w} yields
\begin{equation}
    \frac{\partial \mathcal{L}}{\partial \mathbf{W}} = 
    \frac{\partial \mathcal{L}}{\partial \mathbf{X}^M}
    \left[ \mathbf{I} + \Delta t \left( \mathbf{E}_0  + \sum_{i=1}^M \mathbf{E}_i \right) + \mathcal{O}(\Delta t^2) \right] 
    \frac{\partial \mathbf{X}^0}{\partial \mathbf{W}},
\end{equation}
which reformulates the gradient calculation into a recursive sum. This implies that the gradients will not vanish exponentially, but the gradients may still become very small.
\end{proof}

\section{Other Proofs}\label{appendix:auxiliary_proofs}

We provide additional Lemmas to provide deeper insight into the theoretical properties of our BIMP model.

\begin{tcolorbox}[blanker, interior engine=spartan, colback=CornflowerBlue!25, boxsep=1mm, borderline west={1.5pt}{0pt}{gray}]
\begin{proposition}[Expressive capacity of BIMP] BIMP can model more diverse node feature representations than approaches whose dynamics are equivalent to linear opinion dynamics.
\end{proposition}
\end{tcolorbox}

\begin{proof}
Nonlinear systems can exhibit more complex behavior, such as multiple equilibria and bifurcations, than linear systems. Also, many continuous-depth GNNs \cite{chamberlain2021grand, thorpe2022grand++, choi2023gread, nguyen2024coupled} lack a feature mixing mechanism. In contrast, BIMP introduces the option graph $\mathcal{G}^{\mathrm{o}}$ to enable feature mixing across dimensions, modeling more complex information exchange.

To highlight the contributions of the intrinsic nonlinearity and the option graph $\mathcal{G}^{\mathrm{o}}$, we compare BIMP with linear opinion dynamics as a representative baseline: linear opinion dynamics is a first-order approximation of BIMP without correlated options $\mathbf{A^\mathrm{o}}=\mathbf{0}$.

When attention parameter $u=1$, input parameter $\mathbf{B=0}$, and uncorrelated options $\mathbf{A^\mathrm{o}}=\mathbf{0}$, the BIMP model has dynamics of the form 
\begin{equation}
    \dot{\mathbf{X}}
    = -d\mathbf{X} 
    + \tanh\left( \alpha \mathbf{X} + 
    \mathbf{A}^\mathrm{a} \mathbf{X} 
    \right). 
\end{equation}
When $\mathbf{X=0}$, $\mathbf{\dot{X}=0}$, so $\mathbf{X=0}$ is an equilibrium of the system. The first-order approximation of our model dynamics about this equilibrium is given by
\begin{equation}
    \dot{\mathbf{X}} = (\mathbf{A}^\mathrm{a} - c\mathbf{I}) \mathbf{X}, \quad c=d-\alpha.
\end{equation}
When $c=1$, this equation reduces to,
\begin{equation}
    \dot{\mathbf{X}}
    = ( 
    \mathbf{A}^\mathrm{a} - \mathbf{I} ) \mathbf{X},    
\end{equation}

which is of the same form as linear opinion dynamics in Equation \eqref{eq:FD_model}. Since linear opinion dynamics has the same form as the first-order approximation of BIMP, we say BIMP has greater expressive capacity.
\end{proof}

\begin{tcolorbox}[blanker, interior engine=spartan, colback=CornflowerBlue!25, boxsep=1mm, borderline west={1.5pt}{0pt}{gray}]
\begin{proposition}[Expressive capacity of BIMP can degrade]
    The BIMP model reduces to a linear system when the attention parameter \( u \) is either very small or very large.
\end{proposition}
\end{tcolorbox}

\begin{proof}
The degeneration to a linear model occurs under two settings: (1) when $u$ is very small and the nonlinear term evaluates to 0; (2) or when $u$ is very large such that the hyperbolic tangent saturates, and therefore the nonlinear term evaluates to $\pm 1$.
\end{proof}
To avoid both degenerate cases, we set the attention parameter 
$u$ at the bifurcation point. Beyond the reasoning provided in Section~\ref{sec:u_parameter}, this lemma offers an additional perspective that placing $u$ at the bifurcation point ensures that BIMP operates within the nonlinear regime of Equation~\eqref{eq:BIMP}, thereby preserving its expressive capacity.

\begin{tcolorbox}[blanker, interior engine=spartan, colback=CornflowerBlue!25, boxsep=1mm, borderline west={1.5pt}{0pt}{gray}]
\begin{proposition}[Reduced order representation of BIMP]
\label{lemma:lyapunov_schmit}
    When the input parameter $\mathbf{B}$ is equal to zero, the dynamics of BIMP can be approximated by the dynamics of the reduced one-dimensional dynamical equation.
\begin{equation}
\label{eq:lyapunov_reduction}
    \dot{y} = -d\,y + \operatorname{tanh} \left[ u (\alpha+3) y\right] 
    % + \langle \mathbf{w}_{\max}, \tilde{\mathbf{b}}\rangle
    ,
\end{equation}
where $y = \langle \mathbf{x}, \mathbf{w}_{\max} \rangle \in \mathbb{R}$, and $\mathbf{w}_{\max}$ is the left dominant eigenvector of $\tilde{\mathbf{A}}$.
\end{proposition}
\end{tcolorbox}
\begin{proof}
Leveraging the Lyapunov-Schimit reduction, the BIMP dynamics can be projected onto a one-dimensional critical subspace~\cite{leonard2024fast}. The BIMP dynamics in Equation \eqref{eq:uncorrelated_options} can be vectorized following Lemma \ref{prop:uncorrelated_equation} as
\begin{equation}
    \dot{\mathbf{x}} = -d \mathbf{x}
    + \tanh\left[u \left( (\alpha-1) \mathbf{x} + 
    \tilde{\mathbf{A}} \mathbf{x}
    \right) \right] 
    + \mathbf{b},\label{eq:BINN_vec_appendix}
\end{equation}
where $\mathbf{x} = \operatorname{vec}(\mathbf{X})$, and $\mathbf{b} = \operatorname{vec}(\mathbf{B})$. Defining the right eigenvector matrix $\mathbf{T} = [\mathbf{v}_1, \mathbf{v}_2, \ldots, \mathbf{v}_{N_a}]$ and the left eigenvector matrix $\mathbf{T}^{-1}=[\mathbf{w}_1, \mathbf{w}_2, \ldots, \mathbf{w}_{N_a}]^{\top}$ of $\tilde{\mathbf{A}}$, Equation \eqref{eq:BINN_vec_appendix} can be expressed in the new coordinates $\mathbf{x} = \mathbf{T}\mathbf{y}$ as
\begin{equation}
    \mathbf{T}\dot{\mathbf{y}} = -d \mathbf{T}\dot{\mathbf{y}} + \operatorname{tanh} \left[ u\biggl((\alpha-1)\mathbf{T}\mathbf{y}+\tilde{\mathbf{A}}\mathbf{T}\mathbf{y}\biggr)\right].
\end{equation}
Multiplying $\mathbf{T}^{-1}$ on both sides gives
\begin{equation}\label{eqn:binn_reduced_noapprox}
    \dot{\mathbf{y}} = -d \dot{\mathbf{y}} + \mathbf{T}^{-1} \operatorname{tanh} \left[ u\biggl((\alpha-1)\mathbf{T}\mathbf{y}+\tilde{\mathbf{A}}\mathbf{T}\mathbf{y}\biggr)\right].
\end{equation}

Consider that $c\operatorname{tanh}(x) \approx \operatorname{tanh}(cx)$ for small $|x|$, Equation \eqref{eqn:binn_reduced_noapprox} can be approximated by
% Considering $c\operatorname{tanh}(x) \approx \operatorname{tanh}(cx)$ for small $|x|$, we can put $T^{-1}$ inside $\operatorname{tanh}(\cdot)$:
\begin{equation}\label{eqn:binn_reduced_approx}
    \dot{\mathbf{y}} = -d \dot{\mathbf{y}} + \operatorname{tanh} \left[ u\biggl((\alpha-1)\mathbf{y}+\mathbf{T}^{-1}\tilde{\mathbf{A}}\mathbf{T}\mathbf{y}\biggr)\right] .
    %+ T^{-1} \tilde{\mathbf{b}}
\end{equation}
Defining $\boldsymbol{\Lambda}$ as the diagonal matrix of eigenvalues of $\tilde{\mathbf{A}}$, Equation \eqref{eqn:binn_reduced_approx} can be further simplified by decomposing $\tilde{\mathbf{A}} = \mathbf{T}\boldsymbol{\Lambda}\mathbf{T}^{-1}$
% $\tilde{A}$ can be decomposed as $\tilde{A} = T\Lambda T^{-1}$ where $\Lambda$ is the diagonal matrix of eigenvalues, then the equation further becomes to:
\begin{equation}\label{eq:transfer_binn}
    \dot{\mathbf{y}} = -d \dot{\mathbf{y}} + \operatorname{tanh} \left[ u\biggl((\alpha-1)\mathbf{y}+\boldsymbol{\Lambda}\mathbf{y}\biggr)\right].
    %+ T^{-1} \tilde{\mathbf{b}}
\end{equation}
% Noticing the $\Lambda$ is diagonal matrix, the system can be decomposed into $N_a$ individual sub-systems:
Equation \eqref{eq:transfer_binn} approximates the dynamics of Equation \eqref{eq:BINN_vec_appendix} around $\mathbf{x}=\mathbf{0}$. By observing that $\mathbf{x} = y_1\mathbf{v}_1 + y_2\mathbf{v}_2 + \ldots + y_{N_a}\mathbf{v}_{N_a}$, we can further restrict the dynamics of BIMP to the critical subspace $\operatorname{Ker}(J)=\mathbf{v}_{\max}=\mathbf{v}_1$ through setting $y_2 = y_{3} = \ldots = y_{N_a}=0$. As such, Equation \eqref{eq:transfer_binn} simplifies into 
% Equation \eqref{eq:transfer_binn} is the approximation of equation \eqref{eq:BINN_vec_appendix} around $\tilde{\mathbf{X}}=\mathbf{0}$. Noticing that $\tilde{\mathbf{X}} = y_1\mathbf{v}_1 + y_2\mathbf{v}_2 + ...+ y_{N_a}\mathbf{v}_{N_a}$, we can restrict BINN's dynamics to critical subspace $\operatorname{Ker}(J)=\mathbf{v}_{\max}=\mathbf{v}_1$ by setting $y_2 = ... = y_{N_a}=0$ in equation \eqref{eq:transfer_binn}. Then the equation \eqref{eq:transfer_binn} only keeps 
\begin{equation}
    \dot{y_1} = -d \dot{y_1} + \operatorname{tanh} \left[ u((\alpha-1)y_1+\lambda_1 y_1)\right] .
    %+ \mathbf{w}_i^\top \tilde{\mathbf{b}}
\end{equation}
Substituting $\lambda_1 = \lambda_{\max}^{\tilde{a}}=4$ from Proposition \ref{proposition:lambda} and simplifying $y_{1}$ as $y$ gives
\begin{equation}
    \dot{y} = -d \dot{y} + \operatorname{tanh} \left[ u(\alpha +3y)\right],
    %+ \langle \mathbf{w}_{\max}, \tilde{\mathbf{b}}\rangle
\end{equation}
which we define as the one-dimensional critical subspace for our model. The remaining eigenvectors $\mathbf{v}_{i}$ make up the regular subspace as their eigenvalues are smaller than 0. Systems on the regular subspace vanishes quickly and does not contribute to the long-term behavior (i.e, convergence to equilibrium). It is therefore sufficient to focus on the critical subspace to understand the dynamics of the equilibrium as the regular subspace decays quickly.
\end{proof}

\begin{tcolorbox}[blanker, interior engine=spartan, colback=CornflowerBlue!25, boxsep=1mm, borderline west={1.5pt}{0pt}{gray}]
\begin{proposition}[Formation of consensus in BIMP]
\label{proposition:b=0}
BIMP exhibits oversmoothing when the input parameter $\mathbf{B}$ is equal to zero.
\end{proposition}
\end{tcolorbox}

\begin{proof}
For $\mathbf{x}$ in the neighborhood of the equilibrium $\mathbf{x}=\mathbf{0}$, the Equation \eqref{eq:lyapunov_reduction} in Lemma \ref{lemma:lyapunov_schmit} is isomorphic to
\begin{equation}\label{eq:lyapunov_reduction_iso}
    \dot{y} = (u(\alpha+3)-d)y-u(\alpha+3)y^3.
\end{equation}
At the bifurcation point $u=\frac{d}{\alpha+3}$, Equation \eqref{eq:lyapunov_reduction_iso} has unique equilibrium $y=0$. 

This corresponds to an equilibrium solution of $\mathbf{x} = \mathbf{0}$ in the original system (Equation \eqref{eq:BINN_vec}) which means that $\mathbf{X}=\mathbf{0}$ and all agents form neutral opinions for all options. Since the opinions of all agents have converged, the system has reached consensus (i.e., exhibits oversmoothing). 
\end{proof}

This lemma indicates that BIMP requires an appropriately chosen input term $\mathbf{B}$ to avoid converging to consensus, as discussed in Theorem~\ref{theorem:b!=0}.

%%%%%%%%%%%%%%%%%%%%%%%%%%%%%%%%%%%%%%%%%%%%%

%% file: appendix/2_dataset.tex
%%%%%%%%%%%%%%%%%%%%%%%%%%%%%%%%%%%%%%%%%%%%%%%%%%%%%%%%%%%%%%%%%%%%%%%%%%%%%%%%%%%%%%%%%%%%%%%%%%%%%%%%%
\subsection{Homophilic datasets}
Table~\ref{tab:acc_result_homo} in Section~\ref{subsec:classification} performs semi-supervised node classification task on the Cora~\cite{mccallum2000automating}, citeseer~\cite{sen2008collective}, Pubmed~\cite{namata2012query}, CoauthorCS~\cite{shchur2018pitfalls} and Amazon Computers and Photo~\cite{mcauley2015image} homophilic datasets. While recent continuous-depth models such as GRAND-$\ell$ and KuramotoGNN makes use of the largest connected component of the datasets, we implement classification on full datasets.

\textbf{Cora.}
The Cora dataset contains a citation graph where 2708 computer science publications are connected by 5278 citation edges. Each publication has an 1433-dimensional bag-of-words vector derived from a paper keyword dictionary. Publications are classified into one of 7 classes corresponding to their primary research area.

\textbf{citeseer.}
The citeseer dataset contains a citation graph where 3312 computer science publications are connected by 4552 citation edges. Each publication has a 3703-dimensional bag-of-words vector derived from a paper keyword dictionary. Publications are classified into one of 6 classes corresponding to their primary research area.

\textbf{Pubmed.}
The Pubmed dataset contains a citation graph where 19717 biomedical publications are connected by 44324 citation edges. Each publication is represented by a 500-dimensional TF/IDF weighted word vector derived from a paper keyword dictionary. Publications are classified into one of 3 classes corresponding to their primary research area.

\textbf{CoauthorCS.}
The CoauthorCS dataset is one segment of the Coauthor Graph datasets that contains a co-authorship graph that consist of 18333 authors and connected by 81894 co-authorship edges. Each author is represented by a 6805-dimensional bag-of-words feature vector derived from their paper keywords. Authors are classified into one of 15 classes corresponding to their primary research area.

\textbf{Amazon Computers.}
The Amazon Computers dataset, denoted as Computers in our paper, contains a co-purchase graph where 13381 computer products are connected by 81894 edges. The edges indicate that two products are frequently bought. Each product is represented by a 767-dimensional bag-of-words feature vector derived from their product reviews. Products are classified into one of 10 classes corresponding to their product categories.

\textbf{Amazon Photo.}
The Amazon Photo dataset, denoted as Photo in our paper, contains a co-purchase graph where 7487 photo products are connected by 119043 edges. The edges indicate that two products are frequently bought. Each product is represented by a 745-dimensional bag-of-words feature vector derived from their product reviews. Products are classified into one of 8 classes corresponding to their product categories.
%%%%%%%%%%%%%%%%%%%%%%%%%%%%%%%%%%%%%%%%%%%%%%%%%%%%%%%%%%%%%%%%%%%%%%%%%%%%%%%%%%%%%%%%%%%%%%%%%%%%%%%%%
\subsection{Heterophilic datasets}
Table~\ref{tab:acc_result_hetero} in Section~\ref{subsec:classification} performs semi-supervised node classification task on the Texas, Wisconsin, and Cornell heterophilic datasets from the CMU WebKB~\cite{craven1998learning} project.

\textbf{Texas.}
The Texas dataset contains a webpage graph where 183 web pages are connected by 325 hyperlink edges. Each webpage has a 1703-dimensional bag-of-words vector derived from the contents of the webpage. Webpages are classified into one of 5 classes corresponding to their primary content.

\textbf{Wisconsin.}
The Wisconsin dataset contains a webpage graph where 251 web pages are connected by 512 hyperlink edges. Each webpage has a 1703-dimensional bag-of-words vector derived from the contents of the webpage. Webpages are classified into one of 5 classes corresponding to their primary content.

\textbf{Cornell.}
The Cornell dataset contains a webpage graph where 183 web pages are connected by 298 hyperlink edges. Each webpage has a 1703-dimensional bag-of-words vector derived from the contents of the webpage. Webpages are classified into one of 5 classes corresponding to their primary content.
%%%%%%%%%%%%%%%%%%%%%%%%%%%%%%%%%%%%%%%%%%%%%%%%%%%%%%%%%%%%%%%%%%%%%%%%%%%%%%%%%%%%%%%%%%%%%%%%%%%%%%%%%
\subsection{Large graphs}
Table~\ref{tab:large_graph} in Appendix~\ref{appendix:large_graph} performs semi-supervised node classification task on the ogbn-arXiv~\cite{hu2020open} dataset. The ogbn-arXiv dataset consists of a single graph with 169,343 nodes and 1,166,243 edges where each node represents an arxiv paper, and edges represent citation relationships. We train each model in a semi-supervised way, and compute the training loss over 90,941 of the 169,343 nodes. We use 29,799 of the remaining nodes for validation, and the final 48,603 nodes for testing.

%% file: appendix/3_experiment_details.tex
\subsection{Classification accuracy}\label{experiment:accuracy_appendix}
\subsubsection{Classification accuracy on homophilic dataset}\label{homophilic_appendix}
We demonstrate improved classification performance compared to GRAND-$\ell$, GRAND$++$-$\ell$, KuramotoGNN, GraphCON, GAT, GCN and GraphSAGE. Additionally, we consider oversmoothing mitigation techniques of pairnorm~\cite{zhao2019pairnorm} (denoted -pairnorm) and differentiable group normalization~\cite{zhou2020towards} (denoted -group) and training methods of adaptive attention (denoted -aa) and rewiring with graph diffusion convolution~\cite{atwood2016diffusion} (denoted -rw). Classification performance is reported in Table \ref{tab:acc_result_homo_full} where BIMP outperforms baseline methods.
\begin{table}[h]
    \caption{\textbf{Classification accuracy on homophilic datasets.} Classification accuracies on the Cora, citeseer, Pubmed, CoauthorCS, Computers, and Photo datasets are reported, where BIMP outperforms competitive baselines.}
    \label{tab:acc_result_homo_full}
    \centering
    \setlength\tabcolsep{1.5pt}
        \begin{tabular}{lcccccccc}
        \toprule
        Dataset & Cora & citeseer & Pubmed & CoauthorCS & Computers & Photo \\
        \midrule
        \textbf{BIMP} & {\setlength{\fboxsep}{1pt}\colorbox{LimeGreen!50}{\textbf{83.19$\pm$1.13}}} & {\setlength{\fboxsep}{1pt}\colorbox{LimeGreen!50}{\textbf{71.09$\pm$1.40}}} & {\setlength{\fboxsep}{1pt}\colorbox{Goldenrod!50}{80.16$\pm$2.03}} & {\setlength{\fboxsep}{1pt}\colorbox{LimeGreen!50}{\textbf{92.48$\pm$0.26}}} & {\setlength{\fboxsep}{1pt}\colorbox{Goldenrod!50}{84.73$\pm$0.61}} & {\setlength{\fboxsep}{1pt}\colorbox{Goldenrod!50}{92.90$\pm$0.44}} \\
        \textbf{BIMP-aa} & {\setlength{\fboxsep}{1pt}\colorbox{Goldenrod!50}{82.96$\pm$1.31}} & 70.43$\pm$0.80 & {\setlength{\fboxsep}{1pt}\colorbox{LimeGreen!50}{\textbf{80.35$\pm$0.99}}} & 91.82$\pm$0.37 & 84.72$\pm$0.45 & 92.27$\pm$0.36 \\
        \textbf{BIMP-aa-rw} & 82.59$\pm$1.06 & 70.51$\pm$1.37 & 78.56$\pm$1.12 & {\setlength{\fboxsep}{1pt}\colorbox{Goldenrod!50}{91.97$\pm$0.37}} & {\setlength{\fboxsep}{1pt}\colorbox{LimeGreen!50}{\textbf{84.76$\pm$0.23}}} & {\setlength{\fboxsep}{1pt}\colorbox{LimeGreen!50}{\textbf{92.92$\pm$0.19}}} \\
        GRAND-$\ell$ & 82.20$\pm$1.45 & 69.89$\pm$1.48 & 78.19$\pm$1.88 & 90.23$\pm$0.91 & 82.93$\pm$0.56 & 91.93$\pm$0.39 \\
        GRAND-aa & 82.59$\pm$0.28 & 70.21$\pm$1.21 & 78.39$\pm$1.95 & 91.44$\pm$0.42 & 83.09$\pm$1.71 & 92.50$\pm$0.53 \\
        GRAND-aa-rw & 82.86$\pm$1.47 & {\setlength{\fboxsep}{1pt}\colorbox{Goldenrod!50}{70.95$\pm$1.13}} & 78.56$\pm$1.13 & 91.52$\pm$0.31 & 83.47$\pm$0.51 & 92.64$\pm$0.24 \\
        GRAND++-$\ell$ & 82.83$\pm$1.31 & 70.26$\pm$1.46 & 78.89$\pm$1.96 & 90.10$\pm$0.78 & 82.79$\pm$0.54 & 91.51$\pm$0.41 \\
        GRAND++-aa & 80.14$\pm$0.93 & 69.94$\pm$1.45 & 78.50$\pm$1.28 & 85.65$\pm$1.30 & 84.00$\pm$0.47 & 91.86$\pm$0.52 \\
        GRAND++-aa-rw & 81.91$\pm$1.39 & 69.41$\pm$0.95 & 79.44$\pm$1.06 & 86.23$\pm$0.80 & 83.35$\pm$0.63 & 92.50$\pm$0.22 \\
        KuramotoGNN & 81.16$\pm$1.61 & 70.40$\pm$1.02 & 78.69$\pm$1.91 & 91.05$\pm$0.56 & 80.06$\pm$1.60 & 92.77$\pm$0.42 \\
        GraphCON-Tran & 82.80$\pm$1.34 & 69.60$\pm$1.16 & 78.85$\pm$1.53 & 90.30$\pm$0.74 & 82.76$\pm$0.58 & 91.78$\pm$0.50 \\
        GAT & 79.76$\pm$1.50 & 67.70$\pm$1.63 & 76.88$\pm$2.08 & 89.51$\pm$0.54 & 81.73$\pm$1.89 & 89.12$\pm$1.60 \\
        GCN & 80.76$\pm$2.04 & 67.54$\pm$1.98 & 77.04$\pm$1.78 & 90.98$\pm$0.42 & 82.02$\pm$1.87 & 90.37$\pm$1.38 \\
        GCN-pairnorm & 79.55$\pm$1.21 & 66.93$\pm$0.94 & 76.14$\pm$0.63 & 90.63$\pm$0.69 & 81.88$\pm$2.73 & 86.93$\pm$1.35 \\
        GCN-group & 80.48$\pm$1.40 & 66.99$\pm$1.97 & 77.53$\pm$0.97 & 90.97$\pm$0.54 & 81.97$\pm$0.75 & 89.84$\pm$0.71 \\
        GraphSAGE & 79.37$\pm$1.70 & 67.31$\pm$1.63 & 75.52$\pm$2.19 & 90.62$\pm$0.42 & 76.42$\pm$7.60 & 88.71$\pm$2.68 \\
        \bottomrule
    \end{tabular}
\end{table}

\subsubsection{Classification accuracy on heterophilic dataset}\label{heterophilic_appendix}
We demonstrate improved classification performance compared to GRAND-$\ell$, GRAND$++$-$\ell$, KuramotoGNN, GraphCON, GAT, GCN and GraphSAGE. Additionally, we consider oversmoothing mitigation techniques of pairnorm (denoted -pairnorm) and differentiable group normalization (denoted -group) and training methods of adaptive attention (denoted -aa) and rewiring with graph diffusion convolution (denoted -rw). Classification performance is reported in Table \ref{tab:acc_result_hetero_full} where BIMP outperforms baseline methods.

\begin{table}[h]
    \centering
    \caption{ \textbf{Classification accuracy on heterophilic datasets.} Classification accuracies on the Texas, Wisconsin, and Cornell datasets are reported, where BIMP outperforms competitive baselines.}
    \label{tab:acc_result_hetero_full}
    \setlength\tabcolsep{1.5pt}
    \begin{tabular}{lccc}
        \toprule
        Dataset &  Cornell &  Texas &  Wisconsin \\
        \textit{Homophily level} & 0.30 & 0.11 & 0.21 \\
        \midrule
        \textbf{BIMP} &  {\setlength{\fboxsep}{1pt}\colorbox{Goldenrod!50}{77.13$\pm$3.38}}  & 82.16$\pm$4.06 & {\setlength{\fboxsep}{1pt}\colorbox{LimeGreen!50}{\textbf{86.57$\pm$4.33}}} \\
        \textbf{BIMP-aa} & 76.95$\pm$4.71 & {\setlength{\fboxsep}{1pt}\colorbox{Goldenrod!50}{82.25$\pm$6.49}} & {\setlength{\fboxsep}{1pt}\colorbox{Goldenrod!50}{86.27$\pm$4.36}} \\
        \textbf{BIMP-aa-rw} & {\setlength{\fboxsep}{1pt}\colorbox{LimeGreen!50}{\textbf{77.46$\pm$4.80}}} & {\setlength{\fboxsep}{1pt}\colorbox{LimeGreen!50}{\textbf{82.43$\pm$6.76}}} &  86.22$\pm$4.34 \\
        GRAND-$\ell$ & 70.00$\pm$6.22 &   74.59$\pm$5.43 &  82.75$\pm$3.90\\
        GRAND-aa & 71.89$\pm$5.30 & 75.68$\pm$9.89 & 82.16$\pm$3.77 \\
        GRAND-aa-rw & 74.59$\pm$4.22 & 82.16$\pm$5.43 & 83.73$\pm$5.69 \\
        GRAND++-$\ell$ & 70.30$\pm$8.50 & 76.14$\pm$5.77 & 83.09$\pm$2.83 \\
        GRAND++-aa & 71.89$\pm$3.43 & 78.57$\pm$6.51 & 83.96$\pm$5.02 \\
        GRAND++-aa-rw & 74.05$\pm$5.57 & 79.42$\pm$5.24 & 84.62$\pm$3.56 \\
        KuramotoGNN & 76.02 $\pm$2.77 & 81.81$\pm$4.36 &  85.09$\pm$4.42\\
        GraphCON-GCN &74.05$\pm$3.24   & 80.54$\pm$4.49& 84.79$\pm$2.51  \\
        GAT & 42.16$\pm$7.07 & 57.84$\pm$5.82 & 49.61$\pm$4.21 \\
        GCN & 41.35$\pm$4.69 & 57.03$\pm$5.98 & 48.43$\pm$5.75 \\
        GCN-pairnorm & 52.70$\pm$6.42 & 63.51$\pm$6.54 & 60.59$\pm$4.34 \\
        GCN-group & 47.62$\pm$5.30 & 59.92$\pm$4.15 & 51.37$\pm$5.67 \\
        GraphSAGE & 70.54$\pm$2.55 & 72.70$\pm$5.47 & 73.14$\pm$6.27 \\
        \bottomrule
    \end{tabular}
\end{table}

\subsubsection{Experiment on large graph}\label{appendix:large_graph}
We demonstrate improved classification performance compared to GRAND-$\ell$, KuramotoGNN, GCN and GAT on large graphs. Classification performance is reported in Table \ref{tab:large_graph} where BIMP outperforms baseline methods.
\begin{table}[h]
    \centering
    \caption{\textbf{Classification accuracy on ogbn-arXiv dataset.} Our BIMP model outperforms GRAND-$\ell$, KuramotoGNN, GCN and GAT on the ogbn-arXiv dataset.} 
    \label{tab:large_graph}
    \begin{tabular}{lc}
        \toprule
        Dataset & ogbn-arXiv  \\
        \midrule
        \textbf{BIMP} &  {\setlength{\fboxsep}{1pt}\colorbox{LimeGreen!50}{\textbf{71.04$\pm$0.94}}} \\
        GRAND-$\ell$ & {\setlength{\fboxsep}{1pt}\colorbox{Goldenrod!50}{70.19$\pm$0.43}} \\
        KuramotoGNN & 66.96$\pm$0.25 \\
        GCN & 61.66$\pm$0.32 \\
        GAT & 69.86$\pm$0.59\\
        \bottomrule
    \end{tabular}
\end{table}

\subsubsection{Homophilic dataset hyperparameters}\label{appendix:fine_tune_homo}
We search the hyperparameters using Ray Tune \cite{liaw2018tune} with 1000 random trials for each dataset and final values are shown in Table \ref{tab:fine_tune_homo}. Experiments were run with 100 random splits and each split trained on 100 seeds (seed value for both random split and training are 1-100).
\begin{table}[h]
    \centering
    \caption{\textbf{Hyperparameter for homophilic dataset.} The hyperparameters for homophilic datasets in Section~\ref{subsec:classification} is reported.}
    \label{tab:fine_tune_homo}
    \setlength\tabcolsep{3.5pt}
    \begin{tabular}{lccccccc}
    \toprule
    Dataset   & Cora & citeseer & Pubmed &  CoauthorCS& Computers & Photo\\
    \midrule
    Opinion Dim. & $80$ & $128$ & $128$ & $16$ & $128$ & $64$\\
    Epoch & $100$ & $250$ & $600$ & $250$ & $100$ & $100$\\
    Learning Rate & $0.0178$ & $0.0034$ & $0.0210$ & $0.0018$ & $0.0035$ & $0.0056$\\
    % Regularize & False & False & False & False & False & False\\
    Optimizer & AdaMax & AdaMax & AdaMax & RMSProp & Adam & Adam\\
    Weight Decay & $0.0078$ & $0.1$ & $0.0020$ & $0.0047$ & $0.0077$ & $0.0047$\\
    Dropout & $0.1353$ & $0.3339$ & $0.0932$ & $0.6858$ & $0.0873$ & $0.4650$\\
    Input Dropout & $0.4172$ & $0.5586$ & $0.6106$ & $0.5275$ & $ 0.5973$ & $0.4290$\\
    Attention Head & $4$ & $2$ & $1$ & $4$ & $4$ & $4$\\
    Attention Dim. & $16$ & $8$ & $16$ & $8$ & $64$ & $64$\\
    Attention Type & Scaled Dot & Exp. Kernel & Cosine Sim. & Scaled Dot & Scaled Dot & Pearson\\
    NODE Adjoint & False & False & True & True & True & True\\
    Adjoint Method & n/a & n/a & Euler & dopri5 & dopri5 & dopri5\\
    Adjoint Step Size & n/a & n/a & $1$ & $1$ & $1$ & $1$\\
    Integral Method & dopri5 & dopri5 & dopri5 & dopri5 & dopri5 & dopri5\\
    Linear Encoder & True & True & False & True & True & False\\
    Linear Decoder & True & True & True & True & True & True \\
    Step Size & $1$ & $1$ & $1$ & $1$ & $1$ & $1$\\
    Time ($T$) & $12.2695$ & $6.6067$ & $9.7257$ & $4.0393$ & $3.2490$ & $2.0281$ \\
    Damping ($d$) & $0.8952$ & $1.0970$ & $0.6908$ & $0.1925$ & $1.0269$ & $1.0230$ \\
    Self-reinforce ($\alpha$)  & $1$ & $1$ & $1$ & $1$ & $1$ & $1$ \\
    \bottomrule
    \end{tabular}
\end{table}
\subsubsection{Heterophilic dataset hyperparameters}\label{appendix:fine_tune_hetero}
We search the hyperparameters using Ray Tune with 200 random trials for each dataset and final values are shown in Table \ref{tab:fine_tune_hetero}. Experiments were run with 10 standardized splits and each split trained on 100 seeds (seed value for training are 1-100).
\begin{table}[h]
    \centering
    \caption{\textbf{Hyperparameter for heterophilic dataset.} The hyperparameters for heterophilic datasets in Section~\ref{subsec:classification} is reported.}
    \label{tab:fine_tune_hetero}
    \setlength\tabcolsep{3.5pt}
    \begin{tabular}{lccccccc}
    \toprule
    Dataset   & Texas & Wisconsin & Cornell\\
    \midrule
    Opinion Dim. & $256$ & $32$ & $32$ \\
    Epoch & $200$ & $100$ & $100$ \\
    Learning Rate & $0.0178$ & $0.0178$ & $0.0218$\\
    % Regularize & False & False & False \\
    Optimizer & AdaMax & AdaMax & AdaMax \\
    Weight Decay & $0.0078$ & $0.0091$ & $0.0478$\\
    Dropout & $0.6531$ & $0.2528$ & $0.2030$\\
    Input Dropout & $0.0052$ & $0.0042$ & $0.0417$\\
    Attention Head & $8$ & $4$ & $4$ \\
    Attention Dim. & $32$ & $16$ & $16$ \\
    Attention Type & Scaled Dot & Scaled Dot & Scaled Dot\\
    NODE Adjoint & False & False & False \\
    % Adjoint Method & n/a & n/a & n/a \\
    % Adjoint Step Size & n/a & n/a & $1$ \\
    Integral Method & dopri5 & dopri5 & dopri5 \\
    Linear Encoder & False & False & False \\
    Linear Decoder & True & False & False \\
    Step Size & $1$ & $1$ & $1$ \\
    Time ($T$) & $0.01$ & $0.01$ & $0.01$  \\
    Damping ($d$) & $0.0086$ & $0.0075$ & $0.0195$  \\
    Self-reinforce ($\alpha$)  & $2$ & $2$ & $1.5$ \\
    \bottomrule
    \end{tabular}
\end{table}

\subsection{Performance at large depths}\label{appendix:experiment_depth_1}

\subsubsection{Classification accuracy}\label{appendix:experiment_depth}
Experiment \ref{experiment_depth}: Classification accuracy evaluates the classification performance of BIMP and continuous-depth baselines at different depths of $T=\{1,2,4,8,16,32,64,128\}$ with 100 splits and 10 random seeds. We use the classification accuracy as a measure of the robustness to deep layers of BIMP and baseline methods. 

\begin{figure}[h]
    \centering
    \includegraphics[width=\textwidth]{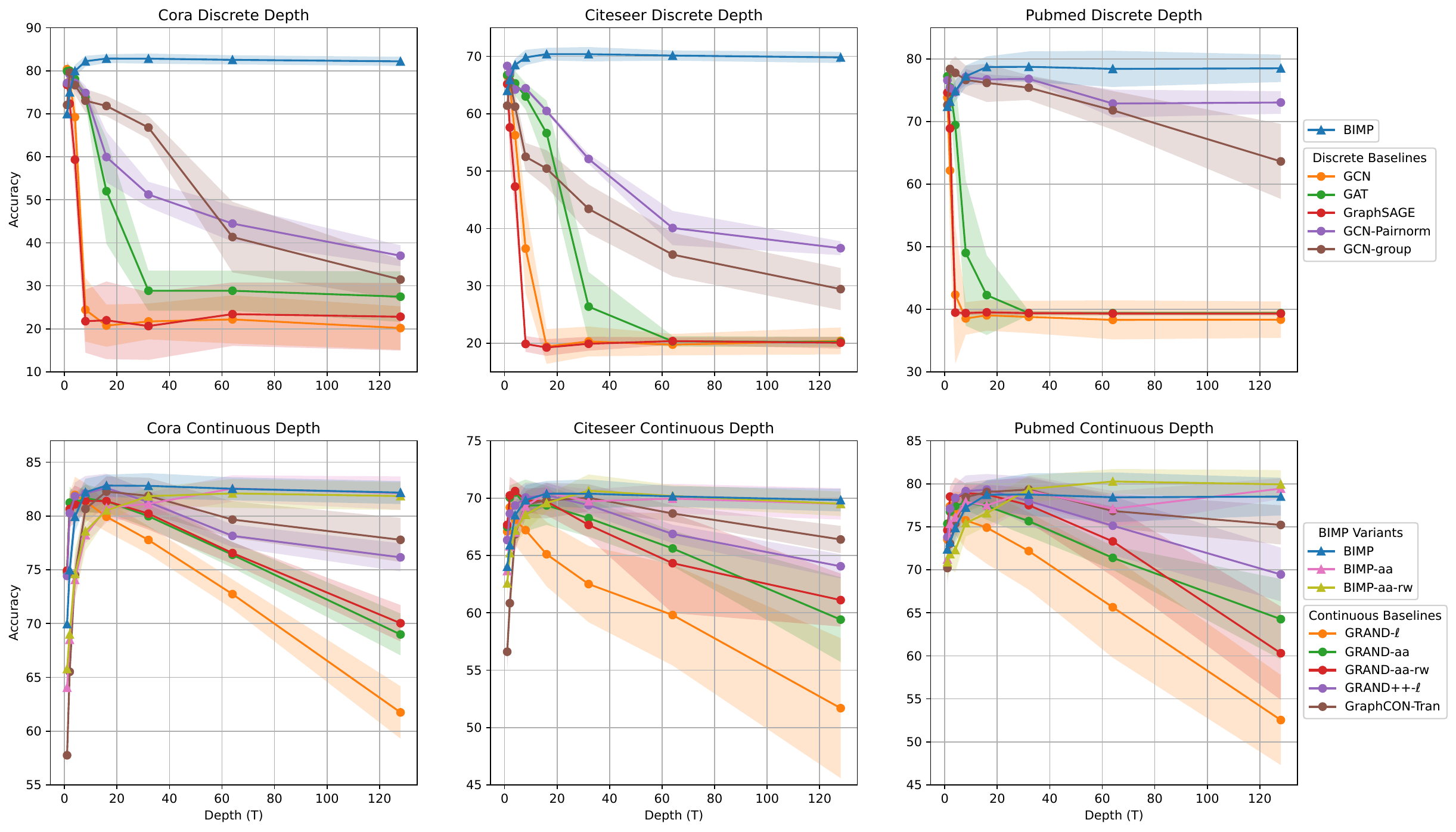}
    \vskip -0.05in
    \caption{\textbf{Classification accuracy.} BIMP is designed to learn node representations that resist oversmoothing even for very large depths. We compare the classification accuracy of BIMP to baseline models for architectures with $1, 2, 4, 8, 16, 32, 64$ and $128$ timesteps. Our BIMP model and its variants are stable out to 128 timesteps, while baseline performance deteriorates after 32 timesteps.}
    \label{fig:accuracy_full}
    \vskip -0.1in
\end{figure}

Figure \ref{fig:depth} show the comparison of classification accuracy of BIMP and select continuous baseline methods. Figure \ref{fig:accuracy_full} show the comparison of classification accuracy of BIMP and additional baselines and oversmoothing mitigation techniques. 

Since adaptive step-size methods like Dormand–Prince (Dopri5) can result in inconsistent numbers of integration steps, we implement the Euler method with fixed step size $\Delta t =1$ for fair comparison. 
Notably, BIMP outperforms GRAND++-$\ell$ at significant depths, even though GRAND++-$\ell$ only supports Dormand–Prince (Dopri5).

Some baselines incorporate an additional learnable weight to scale the differential equation. For instance, in GRAND-$\ell$, the implementation was modified as $\dot{\mathbf{x}} = \alpha [(A-I) \mathbf{x} + \beta \mathbf{x}(0)]$, where $\alpha$ acts as a time-scaling factor. To eliminate its influence and ensure consistency, we set $\alpha =1$ across all methods. 

For each method, we use the fine-tuned parameters provided by each baseline and fix the set of hyperparameters across all depths. For all experiments, we run 100 train/valid/test splits for each dataset with 10 random seeds for each split.

\begin{figure}[h]
    \centering
    \includegraphics[width=\textwidth]{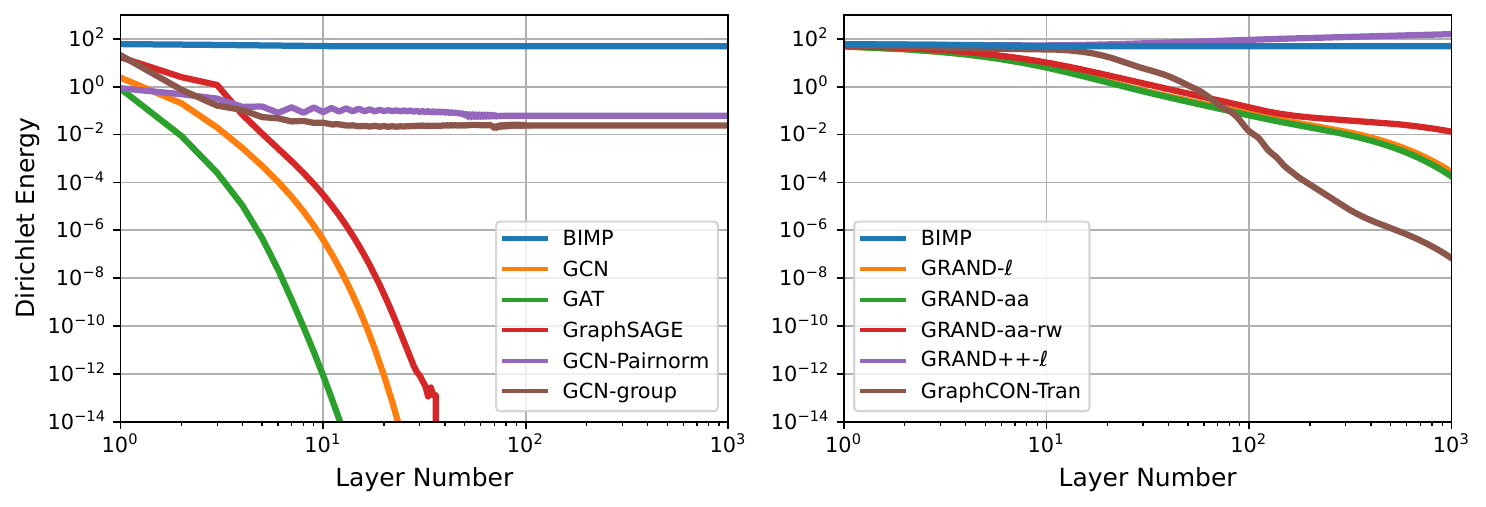}
    \vskip -0.05in
    \caption{\textbf{Dirichlet energy.} BIMP is designed to learn node representations that resist oversmoothing even for very large depths. We compare the Dirichlet energy of node features over a range of network depths. The Dirichlet energy of BIMP remains stable even at very deep layers, while the energy of baseline modes does not.}
    \label{fig:dirichlet_full}
    \vskip -0.1in
\end{figure}

\subsubsection{Dirichlet energy}\label{appendix:experiment_dirichlet}
Experiment \ref{experiment_depth}: Dirichlet energy illustrates the dynamics of the Dirichlet energy in BIMP and baseline methods, which indicates the similarity between the learned features.  
We randomly generate an undirected graph with 10 nodes each with 2-dimensional features sampled from $\mathcal{U}[0,1]$. We randomly initialize the models with the same seed and the node features are propagated forward through 1000 layers. 

Figure \ref{fig:depth} show the comparison of Dirichlet energy of BIMP and select continuous baseline methods. Figure \ref{fig:dirichlet_full} show the comparison of Dirichlet energy of BIMP and additional baselines and oversmoothing mitigation techniques.

\section{Additional Experiments}\label{appendix:additional_experiment}
\subsection{ Computational complexity}
BIMP's space complexity is higher than other baselines as its increase in expressive capacity comes from the introducing of an additional option adjacency matrix $\mathbf{A}^{\mathrm{o}}$. Specially, we compare the space complexity of BIMP with that of GRAND-$\ell$. The complexity of BIMP is $\mathcal{O}(mN_{\mathrm{o}}+N_{\mathrm{o}}^2)$, where $m$ is the number of edges, $N_\mathrm{a}$ is the number of nodes, and $N_\mathrm{o}$ is the number of options. The space complexity of GRAND-$\ell$ is $\mathcal{O}(mN_o)$. Given that the number of options is generally smaller than the number of agents, $N_o$ remains relatively small compared to the number of edge $m$, resulting acceptable computational overhead. 

Additionally, BIMP introduces a nonlinearity through the saturation function $\text{tanh}(\cdot)$, which increases computational cost. However, considering this function operates element-wise, it is more efficient than other nonlinear dynamical processes, such as KuramotoGNN. To illustrate this difference, we compare the average run time of our model against baseline models in Table~\ref{tab:running_time}. We note that BIMP has comparable run time performance to linear baselines models such as GRAND++-$\ell$.

We record the running time for BIMP and other 4 popular continuous-depth GNNs, GRAND-$\ell$, GRAND++-$\ell$, GraphCON-Tran and KuramotoGNN on Cora and citeseer dataset. We train each model 100 times with with a fixed number of epoch (100 for Cora and 250 for citeseer) using fine tuned hyperparameters. The average training time listed in Table \ref{tab:running_time} demonstrates, in contrast to other nonlinear methods like KuramotoGNN, BIMP maintains a training time comparable to other linear continuous-depth baselines.

All experiments reported in the paper was conducted on work stations with an Intel Xeon Gold 5220R 24 core CPU, an Nvidia A6000 GPUs, and 256GB of RAM.

\begin{table}[h]
\centering
\caption{\textbf{Comparable running time.} The average running time (in seconds) for each fine-tuned method tested on the Cora dataset for $100$ epochs and the citeseer dataset for $250$ epochs. Our BIMP model exhibits a modest increase in running time. }
\begin{tabular}{lcccccc}
\toprule
Dataset   & BIMP (ours) & GRAND-$\ell$ & GRAND++-$\ell$ & GraphCON-Tran & KuramotoGNN \\
\midrule
Cora   & 14.33 & 11.97 & 14.06 & 12.86 & 201.32 \\
citeseer   & 42.38 & 31.34 & 41.47 & 15.84 & 252.73 \\
\bottomrule
\end{tabular}
\label{tab:running_time}
\end{table}
\subsection{Choice of nonlinearity in NOD module}\label{appendix:nonlinearity}
To understand how the choice of nonlinearity in our Nonlinear Opinion Dynamics (NOD) module impacts performance, we experiment with a suite of alternative nonlinearities (softsign, arctan, sigmoid, ReLu and GELU) and linearity (linear). Softsign and arctan satisfy the nonlinearity constraint in the NOD definition (i.e., $S(0)=0, S'(0)=1, S''(0)\neq0$), but sigmoid, ReLu and GELU do not. Specifically, sigmoid does not pass through the origin, ReLu is not differentiable, and GELU does not satisfy $S'(0)=1$. Linear refers to the BIMP model without any nonlinearity. We find that using nonlinearities that meet the NOD criteria effectively prevent oversmoothing, while the others do not. We report the classification accuracy of our BIMP model with alternative nonlinearities in the NOD module in Table \ref{table:nonlinearity_transposed}.

\begin{table}[h]
    \centering
    \caption{\textbf{Nonlinear opinion dynamics nonlinearity ablation.} Classification accuracy of our BIMP model on the Cora dataset using various nonlinearities.}
    \setlength\tabcolsep{2pt}
    \label{table:nonlinearity_transposed}
    \begin{tabular}{lcccccccc}
    \toprule
    Layer & tanh & softsign & arctan & sigmoid & ReLu & GELU & linear\\
    \midrule
    1   & {\setlength{\fboxsep}{1pt}\colorbox{Goldenrod!50}{69.96$\pm$1.45}} & 63.25$\pm$1.73 & 64.05$\pm$1.58 & 60.36$\pm$1.39 & 64.17$\pm$1.72 & 62.45$\pm$1.57 & {\setlength{\fboxsep}{1pt}\colorbox{LimeGreen!50}{\textbf{77.52$\pm$1.44}}}\\
    2   & {\setlength{\fboxsep}{1pt}\colorbox{Goldenrod!50}{75.00$\pm$1.50}} & 68.34$\pm$2.03 & 72.62$\pm$2.40 & 63.20$\pm$1.63 & 67.21$\pm$2.18 & 67.06$\pm$2.25 & {\setlength{\fboxsep}{1pt}\colorbox{LimeGreen!50}{\textbf{81.92$\pm$0.85}}}\\
    4   & {\setlength{\fboxsep}{1pt}\colorbox{Goldenrod!50}{79.93$\pm$1.41}} & 72.37$\pm$1.50 & 76.91$\pm$1.82 & 65.54$\pm$1.80 & 73.16$\pm$1.44 & 72.40$\pm$1.60 & {\setlength{\fboxsep}{1pt}\colorbox{LimeGreen!50}{\textbf{82.08$\pm$1.34}}}\\
    8   & {\setlength{\fboxsep}{1pt}\colorbox{LimeGreen!50}{\textbf{82.21$\pm$1.26}}} & 77.05$\pm$1.67 & 79.84$\pm$1.05 & 63.91$\pm$2.13 & 77.76$\pm$1.47 & 77.56$\pm$1.49 & {\setlength{\fboxsep}{1pt}\colorbox{Goldenrod!50}{81.24$\pm$1.48}}\\
    16  & {\setlength{\fboxsep}{1pt}\colorbox{LimeGreen!50}{\textbf{82.83$\pm$1.12}}} & 79.88$\pm$1.82 & {\setlength{\fboxsep}{1pt}\colorbox{Goldenrod!50}{81.81$\pm$1.54}} & 29.55$\pm$1.80 & 81.32$\pm$0.96 & 81.47$\pm$0.63 & 80.45$\pm$1.43\\
    32  & {\setlength{\fboxsep}{1pt}\colorbox{LimeGreen!50}{\textbf{82.81$\pm$1.19}}} & 81.45$\pm$1.48 & {\setlength{\fboxsep}{1pt}\colorbox{Goldenrod!50}{82.48$\pm$1.81}} & 29.92$\pm$1.22 & 82.51$\pm$1.51 & 27.99$\pm$4.22 & 79.99$\pm$1.21\\
    64  & {\setlength{\fboxsep}{1pt}\colorbox{Goldenrod!50}{82.53$\pm$1.07}} & 81.14$\pm$1.65 & {\setlength{\fboxsep}{1pt}\colorbox{LimeGreen!50}{\textbf{82.94$\pm$0.73}}} & 30.72$\pm$1.02 & 76.56$\pm$3.88 & 29.64$\pm$1.82 & 76.74$\pm$1.86\\
    128 & {\setlength{\fboxsep}{1pt}\colorbox{LimeGreen!50}{\textbf{82.18$\pm$1.06}}} & {\setlength{\fboxsep}{1pt}\colorbox{Goldenrod!50}{81.71$\pm$1.37}} & 81.26$\pm$1.93 & 29.37$\pm$2.77 & 71.18$\pm$7.09 & 26.27$\pm$4.89 & 75.44$\pm$0.89\\
    \bottomrule
    \end{tabular}
\end{table}